%% file: main.tex
\documentclass[times,twocolumn,final]{elsarticle}
\usepackage{totcount}
\newtotcounter{citenum}
\def\oldcite{}
\let\oldcite=\bibcite
\def\bibcite{\stepcounter{citenum}\oldcite}

\usepackage{medima}
\usepackage{framed,multirow}

\usepackage{amssymb}
\usepackage{latexsym}

\usepackage{url}
\usepackage[table,usenames,dvipsnames]{xcolor}
\definecolor{r0}{RGB}{103,0,31}
\definecolor{r1}{RGB}{178,24,43}
\definecolor{r2}{RGB}{214,96,77}
\definecolor{r3}{RGB}{244,165,130}
\definecolor{r4}{RGB}{253,219,199}
\definecolor{w0}{RGB}{247,247,247}
\definecolor{b4}{RGB}{209,229,240}
\definecolor{b3}{RGB}{146,197,222}
\definecolor{b2}{RGB}{67,147,195}
\definecolor{b1}{RGB}{33,102,172}
\definecolor{b0}{RGB}{5,48,97}

\usepackage{amsfonts}
\usepackage{multirow}
\usepackage{mathtools}
\usepackage{scrextend}
\usepackage{hyperref}
\usepackage[numbers]{natbib}
\usepackage{url}
\usepackage{balance}
\usepackage[inline]{enumitem}
\usepackage{makecell}
\usepackage[colorinlistoftodos]{todonotes}
\usepackage{pdflscape}
\usepackage{afterpage}
\usepackage{capt-of}

\usepackage{amssymb}
\usepackage{pifont}
\newcommand{\cmark}{\ding{51}}%
\newcommand{\xmark}{\ding{55}}%

\definecolor{newcolor}{rgb}{.8,.349,.1}

\journal{Medical Image Analysis}

\begin{document}

\verso{Xiao \snm{Liu} \textit{et~al.}}

\begin{frontmatter}

\title{Learning Disentangled Representations in the Imaging Domain}%

\author[1]{Xiao \snm{Liu}\corref{cor1}\fnref{fn1}}
\cortext[cor1]{Corresponding author}
\ead{Xiao.Liu@ed.ac.uk}
\author[1]{Pedro \snm{Sanchez}\fnref{fn1}}
\author[1]{Spyridon \snm{Thermos}\fnref{fn1}}
\fntext[fn1]{Equal contribution. \\ \url{https://doi.org/10.1016/j.media.2022.102516}}
\author[1,2]{Alison Q. \snm{O'Neil}}
\author[1,3]{Sotirios A.\snm{Tsaftaris}}

\address[1]{School of Engineering, The University of Edinburgh, Edinburgh EH9 3FG, UK}
\address[2]{Canon Medical Research Europe, Edinburgh EH6 5NP, UK}
\address[3]{The Alan Turing Institute, London NW1 2DB, UK}


\begin{abstract}
Disentangled representation learning has been proposed as an approach to learning general representations even in the absence of, or with limited, supervision. A good general representation can be fine-tuned for new target tasks using modest amounts of data, or used directly in unseen domains achieving remarkable performance in the corresponding task. This alleviation of the data and annotation requirements offers tantalising prospects for applications in computer vision and healthcare. In this tutorial paper, we motivate the need for disentangled representations, revisit key concepts, and describe practical building blocks and criteria for learning such representations. We survey applications in medical imaging emphasising choices made in exemplar key works, and then discuss links to computer vision applications. We conclude by presenting limitations, challenges, and opportunities.
\end{abstract}


\end{frontmatter}


\section{Introduction}
Imagine the need to develop a method to localise the ventricles in Magnetic Resonance Imaging (MRI) and Computerised Tomography (CT) scans of the brain in patients. This method must be robust to any changes in the imaging process, scanner, and noise, as well as to anatomical and pathological variation. The current deep (supervised) learning paradigm indicates that we \textit{must} present to the system as many examples as possible to instill robustness by learning what is unnecessary, or nuisance \citep{Achille2017}, \textit{e.g.}\ the patient being placed at a rotated angle in the scanner, as opposed to what matters, \textit{i.e.}\ the location of the ventricle.
However, collecting and annotating enough data to cover such real-world variation is an unrealistically time-consuming and costly solution. 
\begin{figure}[!ht]
    \centering
    \includegraphics[width=\columnwidth]{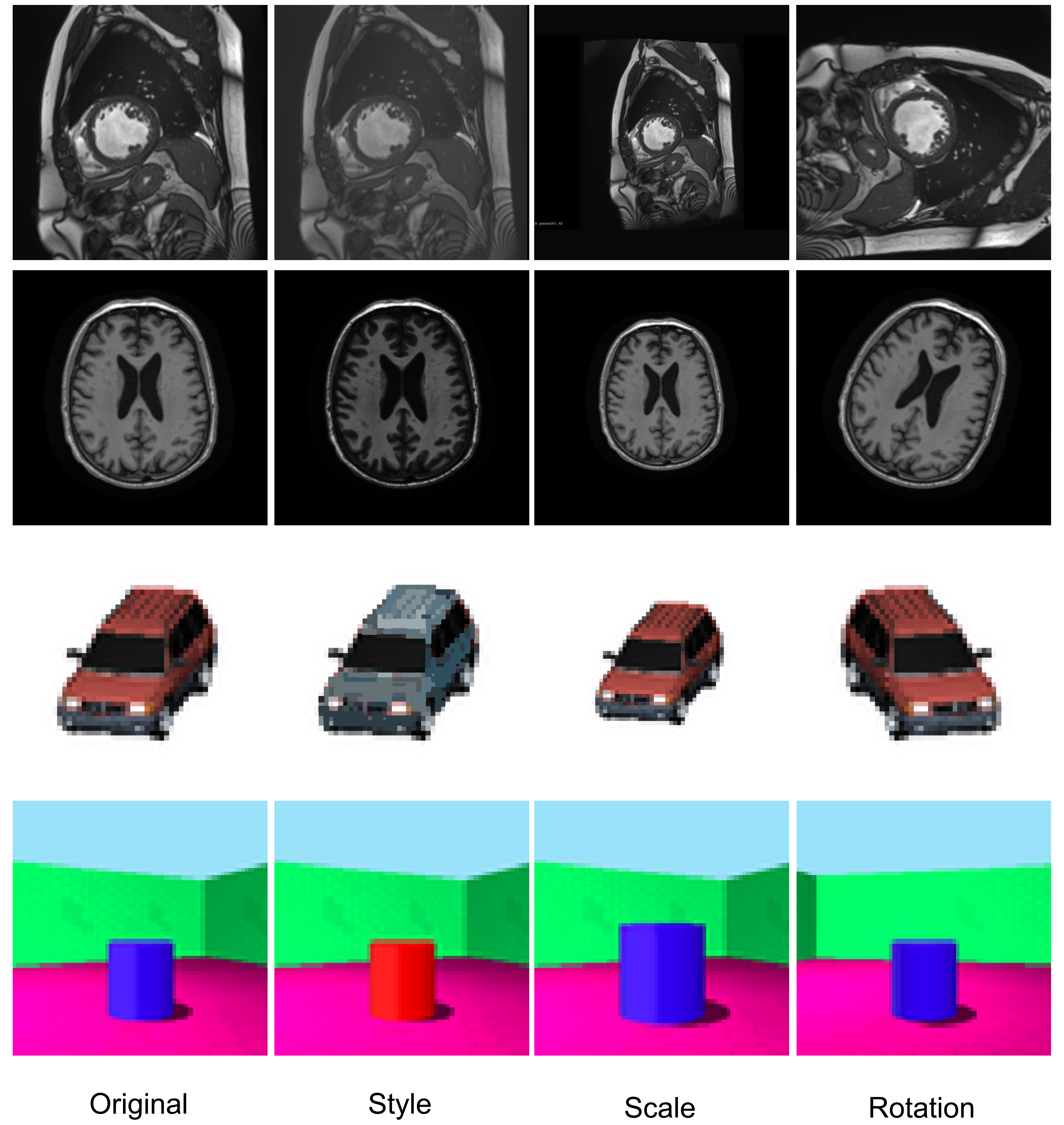}
    \vspace{-2mm}
    \caption{
    Examples of factors of variations: style, scale, and rotation in the context of cardiac scans~\citep{BernardACDC}, brain scans~\citep{petersen2010neurology}, cars \citep{Reed2015neurips}, and 3D shapes \citep{3dshapes18}.}
    \label{fig:intro}
\end{figure}

Surprisingly, we may not always need annotated data or carefully crafted data augmentations to achieve this. With disentangled representation learning (DRL), one learns to encode the underlying factors of variation into separate latent variables \citep{bengio2013pami,higgins2018towards}, which ultimately capture sensitive and useful information for the task at hand and also understand the underlying causal relations amongst the variables. We choose to introduce the reader to DRL by presenting 3 indicative examples of disentangled factors in Fig.~\ref{fig:intro}, which affect the colour, scale, and rotation of the rendered object in the corresponding scene. By adopting DRL, one can design deep models that will be robust to representations from unseen domains, a result that cannot always be achieved through data augmentation.

Herein\footnote{This paper follows a tutorial style, but also surveys a considerable (more than 260 citations) number of works. It aims to be concise and not fully survey the field (in 2021 alone, 1700 papers have appeared on arXiv on disentangled representations). This paper is inspired by two tutorials at MICCAI on disentangled representations (\url{https://vios.science/tutorials/dream2021}) and benefits greatly from feedback received from participants in these tutorials.}, 
we aim to unite understanding of disentangled representation learning and its applications in medical imaging. Our goals are to:
\begin{enumerate}[noitemsep]
    \item Expose why (in)variance matters in learning.
    \item Understand the impact of causal relations in the context of disentangled representations.
    \item Enforce that disentanglement requires at least one of: inductive biases, priors, or supervision. 
    \item Expose building blocks for encouraging disentanglement.
    \item Survey medical image analysis applications.
    \item Draw inspiring lessons from computer vision applications.
    \item Identify limitations, and discuss opportunities and remaining challenges.
\end{enumerate}
To define disentanglement we first revisit key concepts in learning representations. We then provide an overview of key generative frameworks forming the basis of many subsequent models; building blocks of disentanglement; and evaluation metrics. We discuss exemplar models designed to address applications of disentanglement in the medical imaging and computer vision domains. We conclude by discussing opportunities and challenges. This paper is also accompanied by a repository offering links to the implementations of key methods and to existing metrics: \url{https://github.com/vios-s/disentanglement_tutorial}.

%

\section{Revisiting Key Concepts in Representation Learning}
\label{sec:theory}

\textbf{Notation.}~We use $x$, $\mathbf{x}$ and $X$ to denote scalars, vectors, and higher-dimensional tensors respectively, drawn from the domain $\mathcal{X}$ of corresponding dimensions. We use $X_i$ to refer to a datum of the above tensors (of any dimension) for presentation simplicity where tensor dimensionality is implied by the context. We will assume we have access to a dataset containing samples of $X_i$, where $i \in [1,N]$, $N$ denoting the number of samples. We use $\mathcal{X}$ to denote the observed variables of the input domain, $\mathcal{Z}$ for latent representations, $\mathcal{S}$ for real generating factors, and $\mathcal{Y}$ for the output domain. For example, if we choose to solve a classification task, then $\mathcal{Y}$ is a space of scalars $y$.

\subsection{Model learning}
\label{sec:model_learning}
We consider the task of learning a mapping between two domains \citep{vapnik1999overview}, \textit{i.e.}~$f: \mathcal{X} \rightarrow \mathcal{Y}$. We will split $f$ into two components, $f: E_{\phi} \circ D_{\theta}$. $E_{\phi}$ maps to an intermediate latent representation $\mathcal{Z}$ ($E_{\phi}: \mathcal{X} \rightarrow \mathcal{Z}$) whereas $D_{\theta}$ maps to the output ($D_{\theta}: \mathcal{Z} \rightarrow \mathcal{Y}$). We will term $E_{\phi}$ the ``encoder'' and $D_{\theta}$ the ``decoder''.\footnote{$D_{\theta}$ is often referred as a classifier or a regressor, however we avoid this nomenclature here to be more general.} Thus, the goal of model learning is the solution of the task at hand by learning a good representation. Below, we discuss desirable properties of a good representation.

\subsection{Representation learning}
\label{sec:representation_learning}
Finding good representations for the task at hand is fundamental in machine learning \citep{bengio2013pami,scholkopf2021ieee}. Consider the task of detecting brain tumours by placing a bounding box $\mathbf{y}_i$ around each tumour in the image $X_i$. A dataset may contain brain samples with different morphologies, acquired using different protocols in different sites (hospitals), etc. Our goal is to create a representation suitable for the task. If the tumour changes location in the image, we would like the bounding box output to change location accordingly; our representation will be \textit{equivariant} to the location of the object of interest. On the other hand, we would like the representation to be \textit{invariant} to acquisition-related changes.

\textbf{Symmetries.}~Symmetries $\Omega$ are \textit{transformations} that leave some aspects of the input intact \citep{cohen2016icml,Cohen2021thesis,bronstein2021arxiv}. For instance, the category of an object does not change after applying shift operations to the image, therefore these operations are considered symmetries in the object recognition domain. Using the model $f$ and symmetries $\Omega$, we now proceed to define the equivariance and invariance properties.

\textbf{Equivariance.}~A mapping $E_{\phi}: \mathcal{X} \rightarrow \mathcal{Z}$ is equivariant \textit{w.r.t.}~$\Omega$, if there is a transformation $\omega \in \Omega$ of the input $X \in \mathcal{X}$ that affects the output  $Z \in \mathcal{Z}$ in the same manner. Formally, this means that $\Omega$-equivariance of $E_{\phi}$ is obtained when there exists a mapping
$M_\omega: R^d \rightarrow R^d$ applying $\omega$ to an input such that:
\begin{equation}
    E_{\phi}(M_\omega \circ X) = M_\omega \circ E_{\phi}(X), ~\forall \omega \in \Omega.
\end{equation}
In practice, one chooses transformations that induce the desired equivariance and learned properties in accordance with the task at hand, thus a good understanding of the problem (also known as \textit{domain knowledge}) is required \citep{Lenc_2015_CVPR}. Classical examples where equivariance to translation, shift, and mirroring might be important, are image segmentation, pose estimation, and landmark detection tasks.

\textbf{Invariance.}~A special case of equivariance occurs when $M_g$ becomes the identity map. Formally, $E_{\phi}$ is invariant to transformations of $\Omega$ if:
\begin{equation}
    E_{\phi}(M_\omega \circ X) = E_{\phi}(X), ~\forall \omega \in \Omega.
\end{equation}
The transforms we want to adhere to are usually task-specific and as we will highlight in Sec.~\ref{sec:applications} typically enforced via design biases (and costs) to approximate the transformations.

\subsection{Generating factors}
Considering a distribution that characterises the domain $\mathcal{X}$, the \textit{generating factors} $\mathcal{S}$ are the underlying variables that fully characterise the variation of the data --seen or expected to be seen. Recent studies \citep{bengio2013pami, scholkopf2021ieee} argue that representations should enable the decomposition (\textit{i.e.}~disentanglement) of the input data into separate factors. Each factor should correspond to a variable of interest in the underlying process that generated the data. For the rest of the paper we will refer to the real-world generating factors as ``real" and to those learned by a model as ``learned". 

In the brain tumour detection example, several variables such as tumour texture/location, brain shape, acquisition protocol, image contrast, etc.\ may be involved. In general, the more complex the image, the more variables, and the higher the number of possible combinations. Enumerating all these combinations readily leads to a combinatorial explosion in the possible combinations that a dataset must contain to enable a model to learn (from data alone) the desired in/equi-variances. It is not realistic to identify every factor and cover every possible combination. Domain knowledge enables the elucidation of as many factors as possible and allows us to define which real factors we want to be in/equi-variant to. 

\subsection{Domain shifts}
\label{sec:domain_shifts}

An \textit{i.i.d.}~data distribution is easy to consider but forms a strong and often unrealistic assumption. All non-synthetic datasets are somewhat biased due to the finite nature of the acquired data. If learning algorithms are trained with standard supervised learning \citep{vapnik1999overview} without additional assumptions, there is little hope that the learned function will be robust to domain shifts. A model's ability to maintain the desired behaviour across domain changes is also referred to as \textit{out-of-distribution} generalisation \citep{shen2021towards}. For the brain tumour detection example, both CT or MRI scanners acquire images, but we might know that a given hospital uses CT. In this case, modality-related factors are linked to the hospital-related variables. Therefore, understanding the data generation process and the underlying relations between variables can help to distill the important visual information, and to create mechanisms that are more generalisable. Such reasoning enables the design of principled strategies for mitigating the data bias \citep{castro2020causality}. In fact, we can explicitly define the changes we want our model to be invariant or equivariant to, by modeling domain shifts such as: i) population, \textit{i.e.}~different cohorts, ii) acquisition, \textit{i.e.}~different cameras, sites or scanners, and iii) annotation shift, \textit{i.e.}~different annotators.

\subsection{Disentangled representations}
\label{sec:disentangled_representations}
Disentangled representations can address some of the challenges described until now by learning representations with equi/in-variances to specific undesired variables, whilst considering the data generation process and potential domain shifts. Although a widely accepted definition of disentangled representations is yet to be defined, the main intuition is that by disentangling, we separate out the main factors of variation that are present in our data distribution \citep{bengio2013pami,higgins2018towards,caselles2019neurips,locatello2019icml}. We characterise a factor as ``disentangled" when any intervention on this factor results in a specific change in the generated data \citep{caselles2019neurips,thomas2017independently}.

\subsubsection{Formalising disentanglement}
\citet{higgins2018towards} have recently presented a generic definition for disentanglement. Given a compositional world $W$ and a set of transformations $\Omega$ (as defined in Sec.~\ref{sec:representation_learning}), they define a function $f:W\rightarrow Z$ that can induce $\Omega$ in the latent representation $Z \in \mathcal{Z}$ in an equivariant manner. The representation $Z$ is defined as ``disentangled" if there is a decomposition $Z=Z_{1}\times \dots \times Z_{n}$ such that a transformation $\omega$ applied on $Z_{i}$ will result in an equivalent transformation in the input domain $\mathcal{X}$, leaving all other aspects controlled by $Z_{j\neq i}$ unchanged. This definition meets the desired properties of a disentangled representation as defined by several works in DRL \citep{bengio2013pami,chen2016neurips,eastwood2018iclr,ridgeway2018nips}: a) modularity, \textit{i.e.}~each latent dimension should encode no more than one generative factor, and b) informativeness, \textit{i.e.}~all underlying generative factors are encoded in the representation.

A complementary view to the definition of \citet{higgins2018towards} comes from the Information Bottleneck (IB) principle introduced in \citet{tishby1999ccc}. IB allows for learning ``good" representations for the task at hand, by trading-off sufficiency and complexity. Adopting IB, \citet{Achille2017} argue that such representations should be: i) \textit{sufficient} for the task, meaning that we do not discard information required for the output; ii) among all sufficient representations, it should be \textit{minimal} retaining as little information about the input as possible; and finally iii) it should be \textit{invariant} to nuisance effects so that the final classifier will not overfit to any correlations between the dataset nuisances and the ground truth labels.

\begin{figure*}[t]
    \centering
    \includegraphics[width=\textwidth]{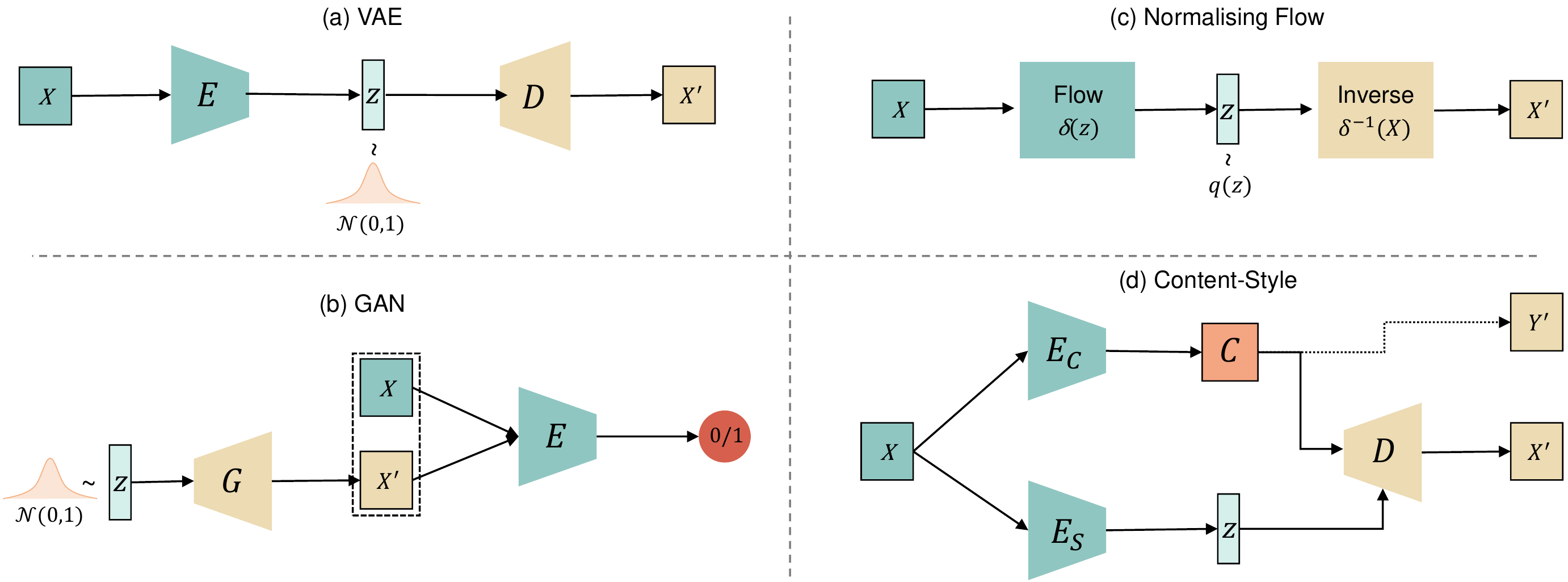}
    \caption{
    Fundamental architectures for disentanglement: a) VAE, b) GAN, c) Normalising Flows, d) Content-Style disentanglement. $X$ and $X'$ are the input and reconstructed images. $\mathbf{z}, C$ are the latent representations, where $C$ represents a tensor latent variable (\textit{e.g.}~image content) and $\mathbf{z}$ represents a vector latent variable. The dashed line in (d) denotes the use of $C$ for learning a representation $Y'$ for a parallel equivariant task (\textit{e.g.}~semantic segmentation). Finally, $\mathcal{N}$ denotes the normal distribution with zero mean and unit variance, whilst $q(\mathbf{z})$ can be any prior distribution.}
    \label{fig:models}
\end{figure*}

\subsubsection{Identifiability}
\label{sec:identifiability}
Learning disentangled representations without any type of supervision is impossible as an infinite family of models that could have generated the observed data exist \citet{locatello2019icml}. Thus, \textit{identifying} the model that generated the data without any additional information is impossible. Given an observation $X_i$, there is an infinite number of generative models that could have generated a sample from the same marginal distribution \citep{locatello2019icml,peters2017elements,Khemakhem2020AISTATS}.

This follows from prior work in non-linear independent component analysis (ICA) \citet{hyvarinen1999nonlinear}: even though the linear case is identifiable, the flexibility given by the non-linear case makes it non-identifiable without extra information.  
Recently \citet{Khemakhem2020AISTATS}, bridged the gap between non-linear ICA and other deep latent variable models, and showed that unsupervised disentanglement methods are, indeed, non-identifiable without additional assumptions.

\subsubsection{A causal perspective}
\label{sec:causality}

Learning disentangled representations of the real factors is not ideal if these factors are not truly independent of each other and are connected via causal relations. Causal relations are directional: the effect will change given an \textit{intervention} (change) on the cause, but not the other way around. For example, the presence of the heart causes the presence of atria, ventricles, pericardium, etc. If an intervention removes the heart, the other structures will also disappear.

Therefore, causal representation learning can be seen as extending DRL \citep{suter2019icml} with additional constraints on the relationships between the latent variables \citep{scholkopf2021ieee,castro2020causality} and potential biases and domain shifts (Sec.~\ref{sec:domain_shifts}). Recent advances in DRL \citep{higgins2017iclr,chen2018neurips,kim2018icml,locatello2019icml} can be cast as learning the causal variables (\textit{i.e.}~the generating factors) of a problem without explicitly modeling the causal mechanisms between them. In addition, identifiability (Sec.\ \ref{sec:identifiability}) can be extended to causality: it is impossible to infer either the latent variables of the generative process or the relationships between them from observational data alone\ \citep{peters2017elements}.

\subsubsection{Disentanglement as inductive bias}
The solution to identifiability is the use of domain knowledge, \textit{i.e.} the \textit{inductive bias}, instead of using explicit supervision \citep{locatello2019icml,peters2017elements,Khemakhem2020AISTATS}. Current representation learning already benefits from the inductive biases of Convolutional Neural Networks (CNNs)~\citep{lecun1998ieee} and Recurrent Neural Networks (RNNs)~\citep{graves2013speech}. Outside of the visual domain, language has been modeled with recurrent neural networks that capture the sequential nature of data for making predictions \citep{lecun2015nature}. Recent attention and self-attention models, such as the transformer architecture \citep{vaswani2017nips}, focus on learning the internal structure of the input data. These self-attention models essentially learn the best inductive biases for each sample in the data distribution.

Overall, disentanglement priors add structure to the learned representations to better correspond to the underlying generation process. It is this useful bias that makes the utilised models identifiable. One of the goals of this paper is to highlight the various inductive biases used. 


\section{Frameworks Enforcing Disentanglement}
\label{sec:frameworks}

We now briefly review fundamental generative models that typically are used to learn disentangled tensor spaces.

\subsection{Variational autoencoders}
\label{sec:vae}
Standard Auto-Encoders (AEs) or Variational Auto-Encoders (VAEs)~\citep{kingma2013auto, rezende2014stochastic} decompose factors via image reconstruction \citep{cheung2014iclrw,siddharth2017neurips}. A typical VAE, depicted in Fig.~\ref{fig:models}(a), discovers and disentangles factors of variation by forcing independence between different dimensions of $\mathbf{z}$, while reconstructing the input $X$. Inter-factor independence is achieved by minimising the Total Correlation (TC) objective imposed on the inferred latent vector \citep{watanabe1960ibm}.

A widely-used VAE that encourages disentanglement is the $\beta$-VAE \citep{higgins2017iclr}. Its main objective is the maximisation of the Evidence Lower-Bound Optimisation (ELBO), ($\mathcal{L}(\theta,\phi;X,\mathbf{z},\beta) = \mathbb{E}_{q_{\phi}(\mathbf{z} \mid X)}[\log  p_{\theta}(X \mid \mathbf{z})] - \beta D_{KL}(q_{\phi}(\mathbf{z} \mid X)||p_{\mathbf{z}})$), to balance (via $\beta > 1$) the reconstruction error versus adherence to the approximate posterior $q_{\mathbf{z} \mid X}$ from the latent prior $p_{\mathbf{z}}$. Note that $p_{\mathbf{z}}$ is usually a normal distribution with identity covariance matrix $\mathcal{N}(0,\mathbf{I})$. The diagonal covariance forces an orthogonal factorisation of the latent space, similarly to a PCA, which reasonably explains the disentanglement capabilities of VAEs \citep{rolinek2019cvpr, burgess2018arxiv, rolinek2019cvpr}. A $\beta > 1$ encourages disentanglement by forcing $q(\mathbf{z} \mid X)$ to carry less information about the reconstruction by increasing the weight of the KL divergence term \citep{burgess2018arxiv} and consequently, increasing independence between the factors of $\mathbf{z}$. Adding more terms such as TC as exploited by several VAE-based models \citep{chen2018neurips, kim2018icml,esmaeili2020aistats} further restricts the redundancy.

\subsection{Generative adversarial networks}
Generative Adversarial Networks (GANs) \citep{goodfellow2014neurips}, see Fig.~\ref{fig:models}(b), typically employ a generator $G$ and a discriminator $D$ in an adversarial game. $G$ generates an image by sampling from an isotropic Gaussian distribution, while $D$ is given the synthetic image and a real one ($X$), and tries to identify which input is real/fake. 

Recent advances in GAN design and training have led to high-fidelity image generation \citep{karras2019cvpr,brock2019iclr,liu2020aaai}. GANs can learn disentangled representations by adding regularisation terms~\citep{chen2016neurips}, by creating an architectural prior \citep{karras2019cvpr}, or even by a post-hoc decomposition of the learned manifold after training \citep{Shen2021CVPR}.

A milestone approach in regularisation is InfoGAN \citep{chen2016neurips} which encourages the disentanglement between two groups of latent variables: a) $\mathbf{z}$ which encodes unstructured noise; and b) $\mathbf{c}$ which captures structured features of the data distribution. They approach this by maximising the mutual information (MI) lower bound between $\mathbf{c}$ and the generated data. Cluster-GAN \citep{mukherjee2019aaai} extends the InfoGAN setting (adopting only the discrete version of $\mathbf{c}$) by employing an inverse-mapping network to project the generated data back to the latent space. This process is supervised by a clustering loss that operates as a regulariser.

Architectural priors were introduced by \citet{karras2019cvpr,Karras2020CVPR}. A mapping network transforms the latent variable $\mathbf{z}$ into intermediate variables that control the style at each convolutional layer of the generator ($G$). Interestingly, this enables feature manipulation at different levels of granularity, \textit{e.g.}~from shape down to texture. This hierarchical structure constitutes arguably a strong prior for disentanglement  \citep{Nie2020icml,peebles2020eccv}. 

\begin{figure*}[t]
    \centering
    \includegraphics[width=\textwidth]{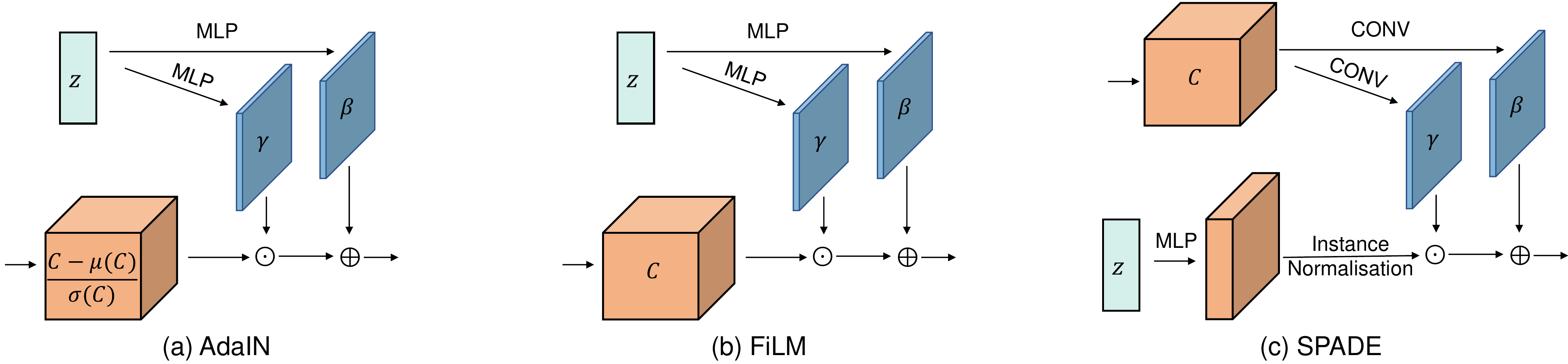}
    \caption{
    Disentanglement building blocks that combine content $C$ with style $\mathbf{z}$: a) AdaIN, b) FiLM, and c) SPADE. $\odot$ and $\bigoplus$ denote element-wise multiplication and addition, respectively. MLP and CONV denote multilayer perceptron and convolutional layers.}
    \label{fig:bblocks}
\end{figure*}

\subsection{Normalising flows}
\label{sec:normalising_flows}
Differently from the non-invertible VAEs and GANs, the Normalising Flows (NFs) are a family of invertible probabilistic models that can compute the exact --and not the approximated as in VAEs-- likelihood \citep{dinh2015iclr,rezende2015icml,2018NEURIPSKingma,papamakarios2021normalizing}. The NF framework derives from the change of variables formula in probability distributions \citep{dinh2015iclr}. Considering a variable $X = \delta^{-1} (\mathbf{z})$, where $\mathbf{z} \sim p(\mathbf{z})$ is sampled from a \textit{prior} distribution $p(\mathbf{z})$, the posterior of $X$ can be obtained by:
    $ p(X) = p(\mathbf{z}) \left| \det~\frac{\partial \delta^{-1}}{\partial \mathbf{z}} \right|^{-1}$.
The \textit{flow} $\delta : \mathbb{R}^d \rightarrow \mathbb{R}^d$ is constrained to be a \textit{diffeomorphism}, \textit{i.e.}~ differentiable and invertible transformations, with a differentiable inverse ($\delta^{-1}$). When multiple flows ($\delta^{-1}_i$) are combined in a chain, they can approximate arbitrarily complex densities for $p(X)$. As Fig.~\ref{fig:models}(c) illustrates, $\delta$ can encode an image $X$ into a latent space $\mathbf{z}$ or using the inverse flow $\delta^{-1}$ to create a generative model by decoding a sample $\mathbf{z} \sim p(\mathbf{z})$ into image space. NFs have been recently adapted to encode disentangled representations \citep{esser2020cvpr,sankar2021glowin} by reinforcing similarity between latent spaces $\mathbf{z}$ of pairs of images with similar generating factors. We refer the reader to \citet{kobyzev2020normalizing} for a comprehensive review.

\subsection{Content-style disentanglement}
\label{sec:cs_disentanglement}

The aforementioned models typically decompose factors into a single vector representation. However, a recent trend in disentanglement focuses on the decomposition of the input image into different latent variables that encode different properties, such as geometry vs.~style. This form of disentanglement is the so-called Content-Style Disentanglement (CSD)~\citep{gatys2016cvpr}, where an image is decomposed into domain-invariant ``content" and domain-specific ``style" representations~\citep{gabbay2020iclr,ruta2021arxiv}. Most works in CSD encode content in spatial (tensor) representations to preserve the spatial correlations and exploit them for a spatially equivariant task, such as Image-to-Image (I2I) translation~\citep{huang2018multimodal,lee2018diverse} and semantic segmentation~\citep{chartsias2019mia}. The corresponding style, \textit{i.e.}~the information that controls the image appearance such as colour and intensity, is encoded in a vector. An abstract visualisation of a CSD model is depicted in Fig.~\ref{fig:models}(d). Note that decomposing content from style is not a trivial process, and encoding content as a high-dimensional representation is not enough. Recent work introduces several design (in terms of the model architecture) and learning (in terms of loss functions) biases to achieve this separation. We denote these inductive biases as ``building blocks" and discuss them in the following section.


\section{Disentanglement Building Blocks}
\label{sec:bblocks}

We now describe common layers and modules that are used at various levels of the model design to encourage disentanglement. We associate these so-called building blocks with different high-level parts of the aforementioned AEs and generative models. We note that typically several of these are combined. In principle we would like to have the minimal set required to solve the task, noting thought that at times these blocks can compete. 

\subsection{Encoding modules}
\label{sec:encoding_modules}
The following are commonly used at various levels of the encoder(s) in popular architectures as bottlenecks. We use representation bottlenecks as a way of reducing the amount of information in the data which will force the network to encode mainly useful concepts.

\textbf{Instance Normalisation.}~Instance Normalisation (IN), originally proposed in~\citep{ulyanov2017instance} for style removal, is commonly used after each convolutional layer of the content encoder to suppress style-related information. In fact, IN removes any contrast-related information from each instance (data sample), encouraging content-related features to be propagated to the following layers. An indicative example is the content encoder of~\citet{huang2017iccv}, where IN replaces all batch normalisation layers~\citep{ioffe2015batch}.

\textbf{Average Pooling.}~ Contrary to IN, average --or global-- pooling is commonly used to suppress the content information in the style encoder~\citep{huang2017iccv}. By averaging values and flattening a spatial feature into a vector, this operator removes any spatial correlation and encodes the global mean statistics (\textit{i.e.}~image style).

\textbf{Parsimony.}~For CSD models that require semantic and parsimonious content for parallel spatially equivariant tasks, there is a need for discretisation of the encoded continuous information. Such discretisation also can help to remove style-related information. The Gumbel Softmax operator is a differentiable solution to this problem. This operator mimics the reparametrisation trick performed in VAEs by sampling from a standard Gumbel distribution and using the Softmax as an approximation of the ``argmax" step that is usually coupled with one-hot operators for discretisation. Another tool that can further restrict the amount of information in a latent space is known as Vector Quantisation (VQ) \citep{van2017neural}. VQ uses a dictionary of learnable entries to restrict the latent features to discrete set of values.

\subsection{Entanglement modules}
Effective recombination or entanglement of the content and style representations in a decoder is vital. The following approaches or layers are commonly used for this purpose at various levels of the decoder in popular CSD architectures. 

\textbf{Concatenation.}~Simple concatenation allows the content and style to be more flexible in capturing the desired information~\citep{lee2018diverse, esser2018cvpr}. However, this may limit the controllability of learning the content and style as the representations may not capture desired information~\textit{e.g.} style representation capturing the shape information.

\textbf{Adaptive Instance Normalisation.}~The Adaptive Instance Normalisation (AdaIN) layer~\citep{huang2017iccv} is commonly used at multiple decoder levels to recombine the content and style representations. As depicted in Fig.~\ref{fig:bblocks}(a), each AdaIN layer performs the following operation: 
\begin{equation}
    \mathrm{AdaIN} = \gamma\frac{C_{j} - \mu(C_{j})}{\sigma(C_{j})} + \beta_{j},
\end{equation}
where each feature map $C_{j}$ is first normalised separately, and then is scaled and shifted based on $\gamma$ and $\beta$, which are parameters of an affine transformation of the style representation (adaptive mean and standard deviation).

\textbf{Feature-wise Linear Modulation.}~As shown in Fig.~\ref{fig:bblocks}(b), Feature-wise Linear Modulation (FiLM)~\citep{perez2017film} is similar to AdaIN. FiLM was initially proposed as a conditioning method for visual reasoning (the task of answering image-related questions). Using FiLM, each channel of the network’s intermediate features $C_j$ is modulated based on $\gamma_j$ and $\beta_j$ as follows: $FiLM(C_j|\gamma_j, \beta_j) = \gamma_j\cdot C_j+\beta_j$, where element-wise multiplication ($\cdot$) and addition are both broadcast over the spatial dimensions. It is used in~\citet{chartsias2019mia} to combine the content and style in the decoder, where $\gamma$ and $\beta$ parameterise the affine transformation of style vectors.

\textbf{Spatially-Adaptive Denormalisation.}~An alternative approach for combining content with style is the use of multiple Spatially-Adaptive Denormalisation (SPADE)~\citep{park2019semantic} layers. As depicted in Fig.~\ref{fig:bblocks}(c), a SPADE block receives the content channels and projects them onto an embedding space using two convolutional layers to produce the modulation parameters (tensors) $\gamma$ and $\beta$. These parameters are then used to scale ($\gamma$) and shift ($\beta$) the normalised activations of the style representation.

\subsection{Encouraging disentanglement in the latent space}
The following operations and priors can be applied on a latent space to encourage disentanglement.

\textbf{Gaussian prior.}~Encouraging the distribution of the encoded (vector) latent representation to match a Gaussian is a common prior. As reported in Sec.~\ref{sec:vae}, such prior encourages the unsupervised disentanglement of the factors of variation and enables sampling for generating new images. 

\textbf{Task priors.}~As discussed in Sec.~\ref{sec:cs_disentanglement}, content representation can be used for a downstream equivariant task, \textit{e.g.}~semantic segmentation. Task losses, such as the segmentation loss, also contribute at learning a disentangled content representation~\citep{chartsias2019mia}. Other task-based priors, \textit{e.g.}~the number of human body parts~\citep{lorenz2019cvpr}, can be leveraged to encourage certain properties for the content.

\textbf{Gradient reversal layer.}~The Gradient Reversal Layer (GRL) was introduced in~\citep{ganin2015unsupervised} for domain adaptation, where the gradient is reversed to prevent the model from predicting undesired results. GRL is effective in learning domain-specific style representations~\citep{gonzalez2018image}. Specifically, when using the style from one domain to generate images with style from another domains, the gradient is reversed to prevent this from happening. 

\textbf{Latent projection.}~Motivated by the findings of \citet{rolinek2019cvpr}, which suggest that VAE encoders cannot model the arbitrary rotations of the representation space, \citet{Zhao2021lssl} propose the projection of the latent space onto the direction with more information about a generating factor. Latent projection allows the information to be disentangled between particular orientations of the data.

\textbf{Frequency Decomposition.}~Recent studies have investigated the use of frequency decomposition transformations to encourage CSD. For example, \citet{liu2021feddg} use the fast Fourier transform to extract image amplitude and phase. Intuitively, the former reflects image style, whereas the latter corresponds to image content. \citet{huang2021fsdr} use Discrete Cosine Transformation (DCT) to extract the domain invariant and domain specific frequency components, as an approximation of content and style factors, respectively.

\textbf{Structured latent.}~A causal approach to representation learning solves the identifiability problem discussed in Sec.~\ref{sec:identifiability} by enforcing the latent space to be structured as a SCM. Structured latents create strong inductive biases because one might not only define the desired variables --which correspond to the generating factors-- but also the relationship between them. This idea can be implemented in different settings by:
\begin{enumerate*}[label=(\roman*)]
    \item using conditional NFs in a VAE latent \citep{pawlowski2020neurips};
    \item  decomposing of a VAE latent space into separate parts, where each component is further processed at different levels of the decoder \citep{leeb2020structured};
    \item constraining the latent variable of a BiGAN \citep{dumoulin2016iclr,donahue2016iclr} with Bayesian networks \citep{dash2020evaluating};
    \item forcing the latent variables of a BiGAN-style architecture \citep{shen2020bidirectional} to follow a graph structure prior defined as an adjacency matrix \citep{shen2020disentangled}.
\end{enumerate*}

\subsection{Learning setups for disentanglement}
Popular learning setups can encourage disentanglement by harmonising the interaction between blocks.

\textbf{Cycle-consistency.}~Cycle-consistency \citep{zhu2017unpaired, almahairi2018icml,hiasa2018cross,zhang2019iclr} is a technique for regularising image translation settings. In particular, it can be useful for reinforcing correspondence between input and generated images~\citep{xia2019consistent,xia2020mia}, or to improve stability and reconstruction fidelity in unsupervised and semi-supervised settings~\citep{li2017neurips}.

\textbf{Latent regression.}~There is a gentle balance to be made in the complexity of these blocks: too complex and with lots of parameter capacity may lead to information captured within their parameters that can lead to this information not captured in the latent variables. Latent regression has been employed to force the reconstructed image to contain information encoded into this representation~\citep{huang2018multimodal}. In particular, considering an input image $X$, the representation $\mathbf{z}$ and the reconstructed image $X'$, we wish to extract a new latent representation $\mathbf{z}'$ from encoding $X'$, which will be as similar as possible to $\mathbf{z}$. In other words, we need to minimise the distance between $\mathbf{z}$ and $\mathbf{z}'$ using a latent regression loss.

\begin{figure*}[t]
    \centering
    \includegraphics[width=\textwidth]{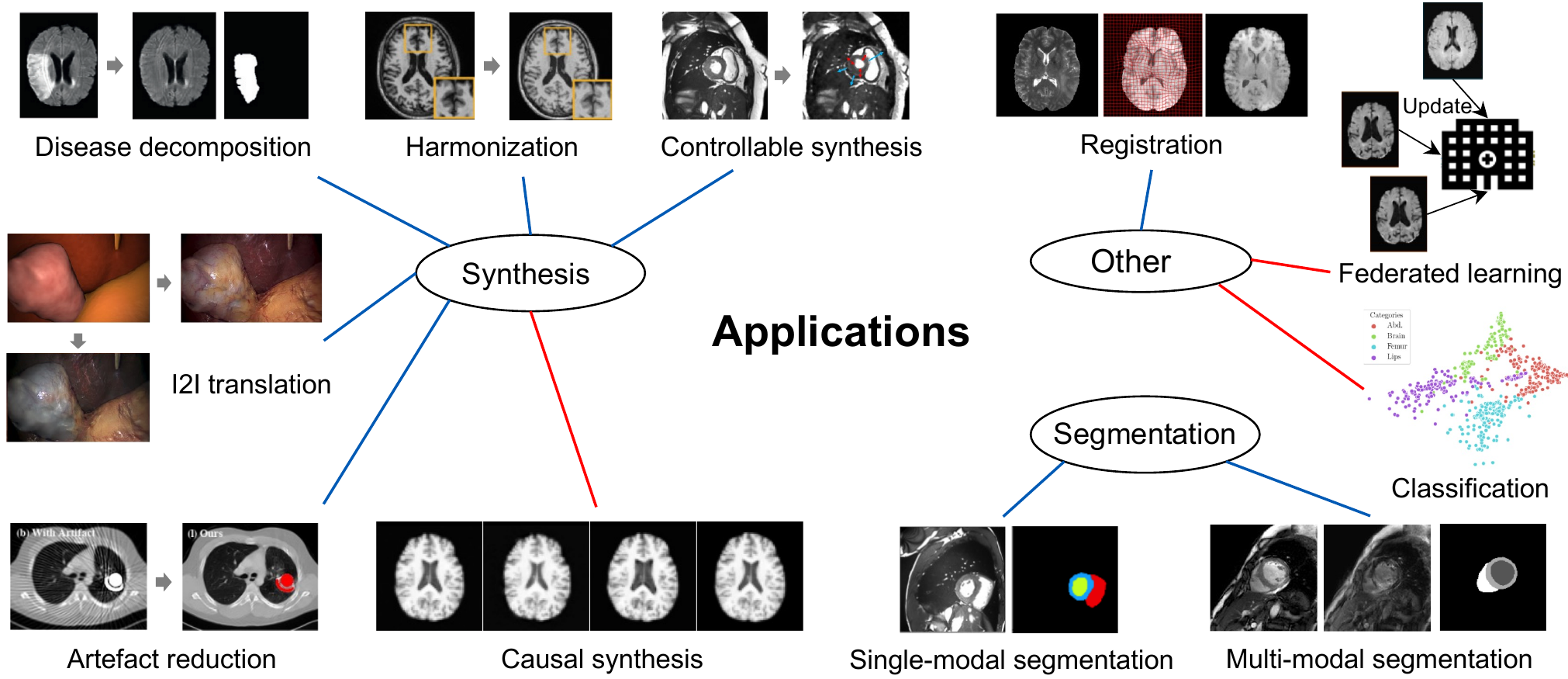}
    \caption{
    Visual summary of disentangled representation learning applications in medical imaging. The visual examples are taken from the exemplar models marked with * in Tab.~\ref{tab:medical_applications_summary}, cited and detailed in Section~\ref{sec:applications}. Red connections indicate vector-based disentanglement, while blue connections indicate tensor/vector-based one (CSD). }
    \label{fig:apps}
\end{figure*}


\section{Metrics for Disentanglement}
\label{sec:metrics}
To understand disentanglement and design models that improve it, we need to be able to quantify how disentangled is (are) the encoded representation(s). Below, we briefly report the most popular disentanglement metrics, splitting them into 2 categories: i) disentanglement of factors in a single vector latent variables, and ii) disentanglement between two latent variables of the same or different dimensionality.

\textbf{Single vector-based latent variable.}~This category consists of both qualitative and quantitative methods for measuring how disentangled a representation is.

Qualitatively, we can evaluate disentanglement by traversing a single latent dimension that alters the reconstructed image by a single aspect (\textit{e.g.}~increase image intensity). In practice, these traversals are linear interpolations which are used to perform ``walks" in non-linear data manifolds and to interpret the variation controlled by each factor \citep{jahanian2020iclr,cherepkov2021cvpr}. Latent traversals do not require ground truth information about the factors. \citet{duan2020iclr} propose a way to quantify latent traversals in a post-hoc fashion, using the unsupervised disentanglement ranking metric to select the most disentangled version of the trained model.
Quantitatively, there has been considerable effort to create metrics to evaluate vector representations. Since there are different proxies for disentanglement, popular metrics focus on measuring different aspects. For example, \citet{higgins2017iclr} propose the first metric to quantify disentanglement when the ground truth factors of a data set are available. In fact, they evaluate disentanglement using the prediction accuracy of a linear classifier that is trained as follows: they first choose a factor $k$ and generate data with this factor fixed, but all others varying randomly. After obtaining the representations of the generated data, they take the absolute value of the pairwise differences of these representations. Then, the mean of these statistics across the pairs gives one training input for the classifier, and the fixed factor index $k$ is the corresponding training output. Subsequently, \citet{kim2018icml} adopt the metric of~\citep{higgins2017iclr}, but construct the training set of the linear classifier by considering the empirical variance of normalised representations rather than the pairwise differences. \citet{chen2018neurips} argue that given a factor of variation, the first two dimensions of the latent vector should have the highest MI. They measure the gap between these two dimensions using the introduced mutual information gap metric. \citet{ridgeway2018nips} propose to measure the modularity of latent representations by measuring the MI between factors, ensuring that each vector dimension encodes at most one factor of variation. \citet{eastwood2018iclr} first train an encoder on a synthetic dataset with predefined factors of variation $z$, and encode a representation $c$ for each data sample. Then, they train a regressor to predict each factor $z$ given a $c$ representation. Based on the prediction accuracy, they measure the disentanglement, completeness, and informativeness of each representation. Finally, \citet{kumar2018variational} propose the separated attribute predictability score to first compute the prediction errors of the two most predictive latent dimensions for each factor, and then use the average error difference as a disentanglement metric. A more comprehensive review of metrics for vector-based disentanglement can be found in~\citet{zaidi2020arxiv}. 

\textbf{Two latent variables.}~The aforementioned metrics are not applicable in CSD as they rely on either having ground truth for the factors or assuming that the latent manifold is solely vector-based. To evaluate CSD one should consider more than one latent variable and a possible difference in dimensionality, \textit{e.g.}~spatial content (tensor) and vector style. To the best of our knowledge, the only work that focuses on CSD metrics is that of \citet{liu2020metrics}. In this work, the authors consider the properties of uncorrelation and informativeness, and propose to combine the empirical distance correlation \citep{szekely2007measuring} and a metric termed information over bias, to measure the degree of disentanglement \textit{between} content and style representations. Two other methods for measuring the uncorrelation-independence between variables of different dimensionality are the kernel-target alignment \citep{cristianini2002kernel} and the Hilbert-Schmidt independence criterion \citep{gretton2005Iica}. However, both methods require pre-defined kernels.


\section{Applications of Disentanglement in Medical Imaging}
\label{sec:applications}

We now survey medical image analysis papers using disentangled representations in a diverse set of challenging tasks. 

\textbf{Strategy.}~We use the search expression ``(disentanglement OR disentangled OR disentangle) AND medical AND (image OR imaging)" over the title and abstract of papers on Google Scholar. 132 papers were found (March 2022) and these were filtered by removing duplicates (with pre-prints) and papers which do not disentangle representations from neural networks.\footnote{We excluded works that used the word ``disentanglement" in context unrelated to representation learning.} After filtering, there were 68 papers utilising disentangled representation in medical imaging.

We categorise these papers based on the investigated application, as shown in the visual summary of Fig.~\ref{fig:apps}. In Tab.\ \ref{tab:medical_applications_summary}, we summarise the survey by highlighting, for each paper, its applications, the general framework for disentanglement, the generating factors being disentangled, the organ over which the method is applied, the imaging modality and we verify if code is available or not.\footnote{All URLs of official implementations for the highlighted methods can be found in the associated repository \url{https://github.com/vios-s/disentanglement_tutorial}.} In the following, for each application, we briefly survey all the relevant papers. We pick one exemplar per application to describe the architecture, training setup, and tips and tricks in their implementation. The choice of the exemplar models is based on how popular the model is, if the code is publicly available, if the model is representative or the first work integrating disentanglement in specific applications and if the model has been extensively evaluated with public datasets.

\input{Tex/applications_summary_table}

\subsection{Synthesis}
\label{sec:image_synthesis}
Many medical imaging procedures are expensive to perform, invasive, and uncomfortable for the patient. For this reason, datasets from certain modalities can be small and imbalanced. Besides, some images are impossible to acquire: doctors might wish to have an image of patient when the patient was healthy in order to perform a comparative diagnosis (these hypothetical image estimations are also called counterfactuals). Otherwise, a training dataset might be acquired in one hospital but deployed in another hospital.  To address these problems, medical image synthesis is considered for augmenting and balancing these datasets. We will now review a few works which utilise disentanglement for synthesising medical images in other to mitigate these issues in different tasks.

\subsubsection{Disease decomposition}
\label{sec:disease_decomposition}
Disease decomposition \citep{xia2019midl,xia2020mia,kobayashi2021decomposing,Tang2021mia,couronne2021longitudinal} aims at disentangling \textit{normal} from \textit{abnormal} factors in an image. \citet{kobayashi2021decomposing} use an architecture inspired on a VAE with vector quantised latent (Sec.\ \ref{sec:encoding_modules}) for disentangling brain tumor from healthy brain information. \citet{Tang2021mia} disentangles normal from abnormal features in lung X-ray using a MUNIT-like~\citep{huang2018multimodal} architecture. Another view is to use temporal information to disentangle disease progression information from changes due to phenotypic differences across subjects \citep{couronne2021longitudinal}. 

We select an image synthesis work of \citet{xia2020mia} because it is extensively validated in public datasets. \citet{xia2020mia} explore disentanglement in the context of synthesising pseudo-healthy brain MRI from patients with tumors or ischemic stroke lesions. The pathology information is disentangled from anatomical features as a segmentation mask.

\textbf{Architecture.}~The model consists of a generator ($G$) extracting a healthy image from a pathological one, and a segmentor ($S$) predicting the remaining pathological information as a mask. Additionally, a decoder module $R$, responsible for reconstructing the input, receives and combines the extracted healthy image and mask enforcing consistency between input and reconstruction. Each module consists of a U-net like architecture with residual blocks and sigmoid output activation, whilst two discriminators follow for the healthy images and the masks.

\textbf{Training setup.}~Two discriminators enforce the (generated) healthy images and the masks to be realistic ($\mathcal{L}_{GAN_1}$ and $\mathcal{L}_{GAN_2}$). When ground truth masks are available, a DICE score is used to learn the segmentations ($\mathcal{L}_{seg}$). Otherwise, adversarial training allows semi-supervised learning of the segmentation with unpaired masks. A cycle-consistency loss ($\mathcal{L}_{CC_1}$) ensures that the subjects keep the ``identity''. The goal is to solely change the pathological aspect while preserving patient identity. Therefore, an additional ($\mathcal{L}_{CC_2}$) cycle consistency objective is introduced to prevent the generator from making unnecessary changes (\textit{e.g.} generation of pathologies for healthy images). The final loss is a combination of the above losses:
\begin{equation}
    \mathcal{L}_{total} = \lambda_1 \mathcal{L}_{GAN_1} + \lambda_2 \mathcal{L}_{GAN_2} +  \lambda_3 \mathcal{L}_{CC_1} +  \lambda_4 \mathcal{L}_{CC_2} +  \lambda_5 \mathcal{L}_{seg}
\end{equation}
The model is evaluated on the following datasets: ISLES \citep{maier2017mia}, BraTS, and Cam-CAN \citep{taylor2017neuroimage} brain datasets.

\textbf{Tips \& Tricks.}~ The main bias introduced in  \citet{xia2020mia} is the use of an auxiliary network for guiding synthesis in a cycle consistency setting. In addition, they observed that, in practice, using a Wasserstein loss coupled with gradient penalty~\citep{gulrajani2017neurips} is more beneficial compared to the Least Squares discriminator loss~\citep{mao2017iccv}.

\subsubsection{Image-to-image translation}
\label{sec:i2i}
Translating one image representation into another, which differs in a specific factor (\textit{e.g.}~style) but maintains others, is termed I2I translation. Translation can be useful in medical imaging when, for instance, one modality is very costly, invasive or even harmful to acquire. In this case, one might choose to acquire an image with a cheaper and/or safer method (with similar content) and subsequently translate the image to the desired domain. MUNIT~\citep{huang2018multimodal} is the first to incorporate a CSD paradigm into I2I translation and it has become a widely adopted benchmark. 
Several others have taken the base architecture from \citet{huang2018multimodal} and extended it several tasks in medical imaging~\citep{kang2019sashimi,Pfeiffer2019miccai}. \citet{kang2019sashimi} translate Fluorescein Fundus (FF) images, which are non-invasive and safe, into Fluorescein Fundus Angiography (FFA), which is the preferred modality for diagnosis but it is invasive and has potential side effects. \citet{Pfeiffer2019miccai} use an I2I architecture for translating synthetic images from a simulator into realistic laparoscopic images for data augmentation. Other works \citep{Zeju2019Anatomy} use a I2I network to disentangle different anatomy factors such lungs and bones in Chest X-ray. \citet{fei2021deep} disentangle content from a modality for synthesising brain MR images of different sequences.

Due to the foundational nature of MUNIT we proceed to detail their learning biases next. 

\textbf{Architecture.}~The basic assumption is that multi-domain images share common content information, but differ in style. A content encoder maps images to multi-channel feature maps, by removing style with IN layers~\citep{ulyanov2017instance}. A second encoder extracts global style information with fully connected layers and average pooling. Finally, style and content representations are combined in the decoder through AdaIN modules~\citep{huang2017iccv}. Disentanglement is further encouraged by a bidirectional reconstruction loss~\citep{zhu2017toward} that enables style transfer. In order to learn a smooth representation manifold, two latent regression losses are applied on content and style extracted from input images, namely a content-based latent regression loss that penalises the distance to the content extracted from reconstructed images, and a style-based latent regression that encourages the encoded style distributions to match their Gaussian prior. Finally, adversarial learning encourages realistic synthetic images. 

\textbf{Training setup.}~MUNIT achieves unsupervised multi-modal I2I translation by minimising the following loss function:
\begin{equation}
    \mathcal{L}_{total} = \mathcal{L}_{GAN} + \lambda_{1}\mathcal{L}_{rec} + \lambda_{2}\mathcal{L}_{c-rec} + \lambda_{3}\mathcal{L}_{s-rec},
\end{equation}
where $\mathcal{L}_{rec}$ is the image reconstruction loss, $\mathcal{L}_{c-rec}$ and $\mathcal{L}_{s-rec}$ denote the content and style reconstruction losses, and $\lambda_{1}=10$, $\lambda_{2}=1$ are the hyperparameters used by the authors. The model has been evaluated on SYNTHIA~\citep{synthia}, Cityscapes~\citep{cityscapes}, edge-to-shoes~\citep{zhu2016generative}, and summer-to-winter~\citep{zhu2017unpaired} datasets. In the medical imaging, \citep{kang2019sashimi} applied a similar strategy to translation from Fluorescein Fundus (FF) towards Fluorescein Fundus Angiography (FFA) images of the eye with the Isfahan MISP~\citep{Alipour2012retinopathy} dataset. \citet{Pfeiffer2019miccai} applied a variation of MUNIT for synthesising realistic laparoscopic images from synthetic images. With multi-modal images of the same patient available, \citet{fei2021deep} achieve better synthesis performance by fusing the shared content information of multi-modal images and creating the pseudo-target modality image as the pseudo-supervision of the synthesised image. 

\textbf{Tips \& Tricks.}~MUNIT has shown robust and impressive performance on multiple I2I scenarios. The style representation is sampled directly from $N(\mathbf{0}, \mathbf{1})$, which means the style latent space is smoother and better for style traversal compared to the style learned by minimising KL divergence using the re-parameterisation trick \citep{kingma2013auto}. Assuming a semantic content prior, the AdaIN layers in the decoder can be replaced with the SPADE module to achieve more controllable translation. The provided hyperparameters can be used for most datasets without the need for intensive tuning. The major drawback of MUNIT is the vague definition of content~\textit{i.e.} the domain-invariant representation, which is achieved by the bi-directional reconstruction. The content is not interpretable, and it is not trivial to measure how domain-invariant it is.

\subsubsection{Artefact reduction}

The presence of artefacts, noise or speckles is common in medical images due to challenges in acquisition. Metal artefacts appear in computed tomography acquisitions, for instance, when a patient carries metallic implants. One might be able to alleviate this issue by disentangling content space from artefact space \citep{Liao2019_miccai, Liao2019tmi, Niu2021Radiation, Huang2020miccai, tang2022generative}, similar to I2I translation in Sec.\ \ref{sec:i2i}. \citet{Huang2020miccai}, for instance, use a similar architecture for speckle noise reduction in optical coherence tomography (OCT) images. \citet{tang2022generative} disentangle noise information from image information with an attention based module for image restoration. The artefact disentanglement network (ADN) \citep{Liao2019_miccai,Liao2019tmi}, reduces the artefacts by disentangling content from artefacts in the latent space utilising unpaired data (\textit{i.e.} unsupervised). \citet{Niu2021Radiation} extends ADN by further reinforcing disentanglement using regularisation in a lower dimensional manifold with ideas from differential geometry. 

We now discuss ADN because their method is extensively validated on publicly available datasets. 

\textbf{Architecture.}~ The architecture in ADN~\citep{Liao2019tmi} contains two groups of encoders and decoders, one for each domain of images with ($\mathcal{I}_a$) and without ($\mathcal{I}$) artefacts. For domain $\mathcal{I}$, an encoder $E_c^{\mathcal{I}}$ outputs a content latent representation $z_c$ with the entire image information and a decoder $D^{\mathcal{I}}$ reconstructs the artefact-free image. For domain $\mathcal{I}_a$, two encoders $E_c^{\mathcal{I}_a}$ and $E_a^{\mathcal{I}_a}$ split the images into two disentangled representations $z_c$ and $z_a$, and a decoder $D^{\mathcal{I}_a}$ reconstructs the image with artefacts. The goal is to learn a transformation from $\mathcal{I}_a \rightarrow \mathcal{I}$ by using $D^{\mathcal{I}} \circ E_c^{\mathcal{I}_a}$ for a given image with artefacts. In addition, one discriminator for each domain is present for reinforcing realism in the reconstructed images.

\textbf{Training setup.}~During training, unpaired images are used as inputs such that the images with artefacts are split into $z_c$ and $z_a$ and the images without artefacts are encoded into $z_c$. By using the decoders to reconstruct versions of the input image both with and without artefacts, they train the neural networks with the following losses:
\begin{equation}
    \mathcal{L}_{total} = \lambda_{1}(\mathcal{L}_{adv}^{\mathcal{I}_a} + \mathcal{L}_{adv}^{\mathcal{I}_a}) + \lambda_{2}\mathcal{L}_{art} + \lambda_{3}\mathcal{L}_{rec} + \lambda_{4}\mathcal{L}_{self},
\end{equation}
where $\mathcal{L}_{adv}$ are adversarial losses, $\mathcal{L}_{rec}$ and $\mathcal{L}_{self}$ are image reconstruction losses computed using cycle consistency, and $\mathcal{L}_{art}$ is a loss that forces the reconstructed image to be anatomically precise.

\textbf{Tips \& Tricks.}~ The artefact decoder takes two latent spaces as inputs (content + artefact representations). \citep{Liao2019tmi} merge these latent spaces using a variation of the feature pyramid network (FPN) \citep{lin2017feature}. The artefact representations  are concatenated at several levels of the artefact encoder with the decoder, using 1x1 convolutional layers to locally merge the features. 

\subsubsection{Harmonisation}

Harmonisation refers to an I2I translation process in medical imaging that aims to reduce domain shifts between acquisitions and improve generalisation \citep{bashyam2020medical}. Some works \citet{Dewey2020miccai,zuo2021informationbased,Zuo2021neuroimage,Wang2021miccaiflow} focus on harmonising magnetic resonance (MR) imaging from different sites. It might even be useful to homogenise acquisitions between different scanners from the same manufacturer \citep{Hongwei2021Uncertainty}. MRI measurements tend to have inconsistencies between images from different sites due to differences in acquisition protocols. \citet{Dewey2020miccai,zuo2021informationbased} disentangle MRI information into content and style information such that input images can be translated to a different style, simulating a different acquisition protocol. \citet{Wang2021miccaiflow} use NF for modelling disentangled domain changes in the latent space of a VAE. 

Next we detail the biases of the method in \citet{Zuo2021neuroimage} because it was extensively validated in publicly available datasets and show superior performance compared to previous methods.

\textbf{Architecture.}~The architecture comprises style $E_s$ and content $E_c$ encoders, and a decoder $D$ that are used for both domains. The harmonisation part of their models utilises 2D slices from the 3D images. This enables to have an intra-patient prior about the style/contrast information: the style should not change within a volume. They also leverage the fact that two MRI sequences (T1 and T2 weighted) are available for each patient, so part of the training can be done in a supervised manner. They also include a discriminator $D_c$ in the content latent space $z_c$ to ensure that the latent is not informative about the style, as opposed to previous methods that used discriminators in image space. The style latent $z_s$ has a probabilistic representation with mean and variance, similar to the VAE in Sec.\ \ref{sec:vae}.

\textbf{Training setup.}~During training, content is disentangled from style in two dimensions: MRI sequence and site. Two images (T1 and T2 sequences) from two different patients (site A and site B) are fed into the network. Therefore, a total of 4 images are used as input. Interestingly, the style encoder $E_s$ and the content encoder $E_c$ have different inputs that belong to the same image. This forces the style latent space $z_s$ to contain only information about the style and not any structural information. Naturally, this setting assumes that different slices in the same image have the same style. The final loss function is
\begin{equation}
    \mathcal{L}_{total} = \lambda_{1}\mathcal{L}_{adv} + \lambda_{2}\mathcal{L}_{z_c} + \lambda_{3}\mathcal{L}_{rec} + \lambda_{4}\mathcal{L}_{percep} + \lambda_{5}\mathcal{L}_{KL},
\end{equation}
where $\mathcal{L}_{adv}$ forces $z_c$ to not contain information about the style, $\mathcal{L}_{z_c}$ encourages the $z_c$ between representations of the same patient but different MRI sequences to be similar, $\mathcal{L}_{rec}$ and $\mathcal{L}_{percep}$, which is a perceptual loss \citep{Johnson2016perceptual}, forces the reconstruction to be the same as the input images.

\textbf{Tips \& Tricks.}~Two other tricks are used for avoiding style information in the content space: the Gumbel-softmax reparameterisation in $z_c$ and random swapping of channels in $z_c$ between latent spaces of the same patient but different MRI sequences.

\subsubsection{Controllable synthesis}

Acquiring annotated data at scale with rare diseases or conditions remains a challenge. It would be extremely useful to have a method that controllably synthesises \citep{liu2018decompose,thermos2021miccai,Liu2020Manifold,Hochberg2021Style,kelkar2022prior,HAVAEI2021mia} images that can correct such underrepresentation. \citet{Hochberg2021Style} uses StyleGAN \citep{karras2019cvpr} with an encoder for controlling style of the synthesised image.  \citet{Liu2020Manifold} and \citet{kelkar2022prior} inject the style from different modalities into the decoder to translate the original image into other styles based on StyleGAN \citep{Karras2020CVPR}. \citet{HAVAEI2021mia} disentangles content from style using conditional GANs and dual adversarial inference \citep{lao2019dual}. \citet{thermos2021miccai} proposed DAAGAN to use the concept of anatomy arithmetic for such controllable generation.

We now discuss in detail DAAGAN due to being first of explicitly doing arithmetic in tensor spaces. 

\textbf{Architecture.}~DAAGAN uses a pre-trained SDNet~\citep{chartsias2019mia} to disentangle the images into anatomy and modality representations. It contains a generator, a pathology classifier, and a discriminator. After extracting the spatial anatomy representations with the pre-trained SDNet, DAAGAN performs the arithmetic operation (mixing and swapping of selected channels) on the anatomy representations of different images (labeled as different pathology or health). The generator takes the mixed novel anatomy factor as input and use AdaIN to combine the anatomy and modality factors to synthesise the corresponding image. In particular, DAAGAN introduced a localised noise injection module in the generator to avoid abrupt mixing of the anatomy channels. The pathology classifier is pre-trained and used to guide the generator to synthesise images with desired pathology. DAAGAN has been evaluated on two cardiac datasets including ACDC~\citep{BernardACDC} and M$\&$Ms~\citep{mnms}.

\textbf{Training setup.}~Apart from the pathology classification loss and the adversarial loss for the generator and discriminator, DAAGAN introduced two consistency losses to encourage the anatomical factors which are not related to the heart to remain unaltered after the arithmetic and noise injection steps. $\mathcal{L}_{cons}$ measures the difference of the background area of the anatomy factor before and after the noise injection and $\mathcal{L}_{bg}$ measures the the difference of the background area of original image and synthesised image.
To train DAAGAN, the total loss is:
\begin{equation}
    \mathcal{L}_{total} = \lambda_{1}\mathcal{L}_{adv} + \lambda_{2}\mathcal{L}_{path} + \lambda_{3}\mathcal{L}_{cons} + \lambda_{4}\mathcal{L}_{bg}.
\end{equation}

\textbf{Tips \& Tricks.}~Since DAAGAN uses SDNet as the extractor of disentangled representations, it is possible to apply DAAGAN to other datasets on which SDNet works well such as CHAOs~\cite{kavur2021chaos} for abdomen, and SCGM~\citep{prados2017spinal} for gray matter and spinal cord. The noise injection module in DAAGAN acts as a mixing corrector to modify the mixed anatomy factors such that the non-suitable mixing of channels from different anatomy factors can be corrected, which has a limitation on the controllability of mixing. Finally, to mix anatomy factors from different images, it is required to first register the anatomy of each image.

\subsubsection{Causal synthesis}

Causal image synthesis \citep{pawlowski2020neurips,reinhold2021arxiv} is a special case of conditional generation in which the conditioning architecture follows a structural causal model (SCM). SCMs consist of graphs where the nodes are generating factors and the edges are causal relationships \citep{peters2017elements}. In fact, the edges represent physical mechanisms of the real world. The representation of each causal variable needs to be disentangled and their relation should be specified by the designer. This is a stronger form of bias than disentangling variables only. With causal models, one might also answer \textit{counterfactual} queries, such as ``What would have happened to an individual if variable `$\mathcal{S}_{i}$' had been different?''. This can be seen as an intervention at individual level. 

CausalGAN \citep{kocaoglu2018causalgan} initially introduced generative models following a causal structure, however, they were not capable of estimating counterfactuals. \citet{pawlowski2020neurips} design a causal model capable of performing counterfactuals with imaging data using normalising flows \citep{rezende2015icml}. A causal graph for a brain MRI problem is constructed where the brain ventricular volume depends on the age, but not on the patient's sex. \citet{reinhold2021arxiv} extend \citet{pawlowski2020neurips} to higher resolution images \citep{dolatabadi2020aistats} and to a more complex SCM of multiple sclerosis disease. \citet{Wang2021miccaiflow} use a similar setup with VAE and NFs for creating a causal model which takes into account image acquisition site information for image harmonisation.

Next we discuss in more detail the method in \citet{pawlowski2020neurips} due to be the first work using generative causal models in medical imaging. 

\textbf{Architecture.}~\citet{pawlowski2020neurips} rely on NFs \citep{rezende2015icml} (invertible neural models (Sec.~ \ref{sec:normalising_flows}) for modeling attributes such as age, sex, brain and ventricular volume and their relationships; and conditional VAEs for synthesising imaging counterfactuals. The conditionals follow a structure based on clinical knowledge about the problem. The authors enable counterfactual estimation by using invertible models. This allows the prediction of a latent representation of an observation and subsequent local intervention by changing the desired latent space.

\textbf{Training setup.}~The networks associated with the NFs and VAEs are trained jointly with backpropagation using the ELBO as a loss function. The model has been evaluated on brain MRI scans from the UK Biobank \citep{sudlow2015plosmed}.

\textbf{Tips \& Tricks.}~The main method used for reinforcing the structural biases is NF, which are based on neural spine flows \citep{durkan2019neurips}. Additionally, the authors realised that normalisation as a pre-processing step is necessary, as it prevents dependencies being learned on the variable with the largest magnitude, and helps with combining the scalar attributes with imaging information. The implementation was done using the Pyro library \citep{bingham2019jmlr} which can be useful for several probabilistic programming tasks.

\subsection{Segmentation}

The goal of deep learning based segmentation is to train a model to accurately predict the pixel-wise labels (segmentation mask) from an image input. Disentangled representation can help by separating out all the information necessary for segmentation (such as the content or shape in an image) from other information such as style.

\subsubsection{Single-modal}
Regarding single-modal medical image segmentation, the input images are acquired with only one modality, \textit{e.g.} MRI images. Spatial Decomposition Network (SDNet)~\citep{chartsias2019mia}, decomposes 2D medical images into spatial anatomical factors (content) and non-spatial modality factors (style). When temporal information is available, temporal consistency objectives can be applied to boost the performance as in~\citep{valvano2019temporal}. Based on SDNet, \citet{jiang2020semi} additionally disentangle the pathology factor to perform semi-supervised pathology segmentation. Disentanglement methods in segmentation also provides the possibility to handle the domain shifts across different domains Additionally, the variational encoding of the style representation allows for sampling and interpolation of the appearance factors, enabling the synthesis of new plausible images~\citep{liu2020mnms}. To learn generalisable representations, gradient-based meta-learning can be applied as a leaning strategy when giving multi-domain data~\citep{liu2021miccai}. \citet{shin2021unsupervised} disentangle intensity and non-intensity for domain adaptation in CT images. \citet{kalkhof2022disentanglement} also disentangles content from style information using a conditional GAN for cross-domain segmentation.

We now detail SDNet as it has been widely used and extended to many medical tasks.

\textbf{Architecture.}~SDNet uses two different encoders for factorising content into a spatial representation and style into a vector one. A decoder is responsible for reconstructing the input by combining the two latent variables, while a segmentation module is applied on the content latent space to learn to predict the segmentation mask for each cardiac part. SDNet learns the content which is represented as multi-channel binary maps of the same resolution as the input. This is obtained with a softmax and a thresholding function. To encourage the style encoder to encode only style-related information, the authors employ a VAE network. Then, style and content are combined to reconstruct the input image by applying a series of convolutional layers with FiLM layers~\citep{perez2017film}. 

\textbf{Training setup.}~SDNet is trained by minimising the following loss function: 
\begin{equation}
    \mathcal{L}_{total} = \lambda_{1} \mathcal{L}_{KL} + \lambda_{2} \mathcal{L}_{seg} + \lambda_{3} \mathcal{L}_{rec} + \lambda_{4} \mathcal{L}_{z_{rec}},
\end{equation}
where $\mathcal{L}_{KL}$ is the KL Divergence measured between the sampled and the predicted style vectors, $\mathcal{L}_{rec}$ is the image reconstruction loss, $\mathcal{L}_{seg}$ is the segmentation loss, and $\mathcal{L}_{z_{rec}}$ is the latent regression loss between the sampled and the re-encoded style vectors. $\lambda_{1}=0.01, \lambda_{2}=10, \lambda_{3}=1,$ and $\lambda_{4}=1$ are the hyperparameters used by the authors. SDNet has been extensively evaluated on the ACDC~\citep{BernardACDC}, MM-WHS~\citep{zhuang2016mia, zhuang2013challenges, zhuang2010tmi}, CHAOs~\cite{kavur2021chaos}, and M$\&$Ms~\citep{mnms} cardiac datasets, as well as on the SCGM~\citep{prados2017spinal} spinal one.

\textbf{Tips \& Tricks.}~SDNet encodes highly semantic content representation, which shows the advantage of content interpretability. 
Other modifications include using a SPADE module to replace FiLM, which has led to a performance improvement in several studies~\citep{liu2020metrics}. Gumbel-Softmax~\citep{jang2016categorical} can replace the naive softmax~\citep{thermos2021miccai} and binary thresholding.

\subsubsection{Multi-modal and cross-modal}
For multi-modal or cross-modal medical image segmentation, at least two modalities are required (\textit{e.g.}~CT and MRI scans). The goal is to accurately predict the segmentation mask given a specific patient, exploiting both (all) available modalities. The most popular models for this task are the Multimodal Unsupervised Image-to-image Translation (MUNIT)~\citep{huang2018multimodal} and the concurrent and similar work of MUNIT, DRIT~\citep{lee2018diverse}. Apart from MUNIT or DRIT, other works include the use of CT data to improve segmentation performance on cone beam computed tomography scans~\citep{lyu2020joint} evaluated on CBCT and CT data~\citep{glocker2013vertebrae}. Similarly, \citet{Runze2021Cross} used CSD for segmentation in a cross-modality setting. SDNet has been also extended to multi-modal setting with the exploitation of aligned Cine and LGE data~\citep{chartsias2019tmi}. \citet{xie2020mi, chen2021beyond} propose cycle-consistency-based GANs to generate better cross-modal images for segmentation by applying mutual information constraints to preserve the image-object information in the content features.

Next we detail how the MUNIT-based models are designed for multi-modal and cross-model medical image segmentation due to the popularity of MUNIT. We refer the readers to Sec.~\ref{sec:i2i} for the details about the architecture and training setup of MUNIT.

\textbf{Architecture.}~In multi-modal or cross-modal medical image segmentation, there are two ways of using MUNIT. The first strategy is aligning the content spaces of data from different modalities~\textit{e.g.} CT and MRI~\citep{yang2020cross, yang2019unsupervised, pei2021disentangle, chen2019robust, ouyang2021representation, ning2021new}. Then, a segmentation network is used to predict the mask with content representations as input. Alternatively, the scenarios of imbalanced domains and domain adaptation is considered, where there is a domain with more data and annotations (\textit{e.g.}~MRI) and a domain with less data and few or no annotations (\textit{e.g.}~CT)~\citep{chen2019unsupervised, jiang2020unified, jiang2022disentangled, liu2022multi, chen2021diverse, wang2022cycmis}. After training a MUNIT model on the two domains, the mappings between them can be obtained. During inference, the samples from CT domain are initially translated to MRI scans, which are then used to predict the segmentation mask by a segmentor trained on the MRI domain. 

\textbf{Training setup.}~MUNIT-based models have been evaluated on the following medical datasets: LiTS~\citep{christ2017lits} liver, NCANDA~\citep{zhao2021association} and BraTS~\citep{menze2014multimodal} brain, CHAOs~\citep{kavur2021chaos} abdomen organs, and MS-CMRSeg~\citep{zhuang2016multivariate, zhuang2018multivariate} and MM-WHS~\citep{zhuang2016mia, zhuang2013challenges, zhuang2010tmi} cardiac datasets, respectively.

\textbf{Tips \& Tricks.}~First, following the typical MUNIT training setup presented in Sec~\ref{sec:i2i}, a consistency loss can be further applied as a regulariser on the task output. For example, the predicted segmentation masks of the original MRI scan and the corresponding translated CT-style images must be consistent~\citep{hoffman2018cycada}. Further, the anatomical (content) latent variables of the different modalities can be aligned or fused by applying adversarial training~\citep{jiang2020unified} and prior constraints as in~\citep{chartsias2019tmi, chen2019robust, ouyang2021representation}. Compared to SDNet, MUNIT's disentangled content is less interpretable. MUNIT needs to be trained using a bi-directional setup. This means that it cannot be used for single-modal datasets.

\subsection{Classification}
A classification task and domain knowledge can be used to disentangle both the task-specific representation $\mathbf{z}_c$ from the classifier and a task-agnostic representation $\mathbf{z}_a$ \citep{benvohen2019embc, meng2019pippi, meng2021tmi, Gyawali2019Ensembling,  Zhao2019miccai, berenguer2020arxiv, harada2021order, zhou2021chest, yang2021disentangled, zhou2022lung}. By merging and decoding the representations, the image can be reconstructed. \citet{berenguer2020arxiv} train a conditional VAE to pre-train an encoder for the subsequent diagnosis classification task. \citet{zhou2021chest, zhou2022lung} disentangle structure and texture on chest X-ray images and show that a pre-trained texture encoder can be efficiently fine-tuned for COVID-19 outcome prediction. \citet{Zhao2021lssl} learn a representation in which a projection is disentangled and \citet{jung2020icaps} use capsules, both resulting in better representations for a downstream classification task. \citet{zhao2021miccai} extend \citet{Zhao2021lssl} to a disentangled direction for different MRI sequences. \citet{yang2021miccai} disentangle time-variant and time-invariant information in longitudinal studies for improving classification based on time-invariant representation. \citet{wang2021miccai} use a graph convolutional AE to disentangle disease-specific and disease-invariant features for improving disease prediction. \citet{harada2021order} disentangle the location-dependent and ulcerative colitis (UC)-dependent representations with the classification losses to achieve semi-supervised learning method for UC classiﬁcation. \citet{bass2021icamreg} detect salient features for classification and regression by explicitly disentangling task-specific and task-agnostic information using ICAM \citep{bass2020icam}. \citet{Cheng2021Clustering} use disentanglement for clustering patients with characteristic phenotypes in order to understand disease progression. \citet{Zou2021interpretation} uses a VAE over meshes for disentangling sex information from hip bones. \citet{Puyol2020Resynchronisation} uses a VAE for disentangling different biomarkers from segmentation masks which are connected to a classifier for interpretability.

We use as exemplar the Mutual Information-based Disentangled Neural Networks (MIDNet) \citep{meng2021tmi}, which was initially developed for ultrasound fetal imaging. Whilst building on earlier work \citep{meng2019pippi}, this approach leverages components that can be easily adapted to other applications and offers a multi-task framework to disentangled task-specific representations. Note that the main goal of disentanglement is to find representations that are invariant to different tasks and domains.

\textbf{Architecture.}~The neural network is composed by two encoders $E_c$ and $E_a$, a classifier $C$ that takes $\mathbf{z}_c$ as input and output the desired class, and a decoder to reconstruct the images by re-entangling the latent spaces $\mathbf{z}_c$ and $\mathbf{z}_a$.

\textbf{Training setup.}~ In addition to classification $\mathcal{L}_{cls}$ and reconstruction $\mathcal{L}_{rec}$ losses $\mathbf{z}_c$ and $\mathbf{z}_a$ are disentangled using Mutual Information Neural Estimation (MINE) \citep{belghazi2018pmlr} ($\mathcal{L}_{MI}$). Domain invariance in $\mathbf{z}_c$ is further reinforced via a clustering loss $\mathcal{L}_{clus}$ that encourages samples from different domains, but with the same label, to have similar task-specific representations. Finally, the network is trained in a semi-supervised way ($\mathcal{L}_{SSL}$) using the MixMatch method \citep{Berthelot2019NEURIPS} and an alignment loss for improving generalisation in a domain adaptation setting. The total loss is:
\begin{equation}
    \mathcal{L}_{total} = \lambda_1 \mathcal{L}_{rec} + \lambda_2 \mathcal{L}_{cls} +  \lambda_3 \mathcal{L}_{MI} +  \lambda_4 \mathcal{L}_{clus} +  \lambda_5 \mathcal{L}_{ssl}.
\end{equation}
The method is evaluated using fetal ultrasound datasets from the iFIND project dataset\footnote{http://www.ifindproject.com/}.

\textbf{Tips \& Tricks.}~
The main inductive bias in \citet{meng2021tmi} is the definition of class specific and class agnostic representations. In addition, the authors use MINE~\citep{belghazi2018pmlr}, which is a learned loss function, for disentangling the two vectors. Other differentiable metrics such as the Hilbert-Schmidt Independence Criterion \citep{gretton2005Iica} could also be used as done in \citet{liu2021miccai}.

\subsection{Registration}
Image registration, defined as the alignment of the content of two images based on a transformation, constitutes an important pre-processing step in medical image analysis. This transformation can be parameterised by either an affine matrix (rigid) or by a displacement field (non-rigid). A major challenge is to define a cost function for multi-modal cases; for example, comparing MRI and ultrasound scans using pixel-level metrics is not effective since the intensity, view and artefacts are different. To address this problem, \citet{chartsias2019tmi} use the disentangled anatomical factors to register the cine- with the LGE-MRI scans of the same patient. The work of \citet{qin2019ipmi} addresses it by leveraging CSD (Sec.\ \ref{sec:cs_disentanglement}). Registration of images with pathologies to atlases can also be problematic, therefore, \citet{Xu2020Reconstruction} disentangles the disease features from the normal features, as in Sec.\ \ref{sec:disease_decomposition}, and generates a deformation field based on the healthy features. We also refer the readers to a recent method~\citep{maillard2022deep} that introduces a deep residual learning implementation of metamorphosis model to handle pathological medical images.

We detail the work of \citet{qin2019ipmi} which uses a CNN to estimate the displacement field from the content space of two images (from different modalities). We choose to discuss this work as it is the first paper to leverage disentanglement for the task of medical image registration.

\textbf{Architecture.}~Initially, a system based on the DRIT~\citep{lee2018diverse} architecture is used for image-to-image translation between two images of different domains. Then, they used the content $E_c$ and style $E_s$ encoders for each domain $\mathcal{X}_{i}$ . Secondly, the content representations from two images are fed into a registration network $G_{reg}$ that outputs deformation fields mapping one image to the other. 

\textbf{Training setup.}~A first training of the CSD is done as defined in \citet{lee2018diverse}. Then, the registration network is learned by computing a bidirectional loss function based on the content latent space of the deformed images plus a regularisation loss over the latent space. The method is evaluated on lung CT scans from the COPDGene \citep{bakas2017advancing} dataset and brain MRI from the BraTS corpus.

\textbf{Tips \& Tricks.}~The main inductive bias used by this model is the fact that registration depends only on the image content; the style can be ignored. In addition, preservation of topological information is an important constraint in medical image registration. The authors use a Huber loss over the gradients of the deformation field for this purpose, reinforcing smoothness of the deformation field.

\subsection{Federated learning}
DRL has just started to be leveraged for maintaining privacy by becoming invariant to private features~\citep{marx2019neurips,Aloufi2020sigsac,bercea2021feddis}. 
Considering that privacy issues in machine learning have attracted significant attention~\citep{liu2020have, su2021patient, jegorova2021arxiv, hartley2022measuring}, we believe that there is a new, emerging domain for learning privacy-preserved disentangled representations. As for every new domain, it will be challenging to connect and exploit existing concepts, such as differential privacy~\citep{dwork2006differential} and federated learning~\citep{rieke2020future, li2020federated, liu2021feddg, bercea2021federated, bercea2021feddis}, with the disentanglement paradigm. Federated learning allows the model to be trained collaboratively by multiple local parties without exchanging or sharing their local data. In this case, the data distributed in local parties are better protected as they are not exposed to external authorities. Among the previous work in federated learning, methods with disentanglement~\citep{liu2021feddg, bercea2021federated, bercea2021feddis} have shown improved performance on Cardiac-CT datasets including CT20~\citep{xu2019whole}, CT34LC and CT34MC~\citep{zhuang2016multi} and several brain MRI datasets.

In particular, the work of \citep{bercea2021feddis}, termed FedDis, learns disentangled representations in the federated learning setting to detect brain anomalies by only sharing the disentangled shape information between clients, while the disentangled appearance information is kept locally. We detail FedDis as it has been extensively evaluated on many public datasets.

\textbf{Architecture.}~FedDis does not have strong design biases to enforce disentanglement, which is mainly achieved by learning biases. FedDis has two auto-encoders to reconstruct the input image and encode the appearance and shape information. 

\textbf{Training setup.}~Apart from the reconstruction loss $\mathcal{L}_{rec}$ for the auto-encoders, FedDis introduces two losses as learning biases to enforce the auto-encoders to separately encode the shape and appearance information. The shape consistency loss $\mathcal{L}_{SCL}$ penalises the difference between the encoded shape embeddings of the original image and the Gamma-shifted image. The latent orthogonality loss $\mathcal{L}_{LOL}$ pushes away the distributions of the shape and appearance embeddings. The overall loss is:
\begin{equation}
    \mathcal{L}_{total} = \lambda_1 \mathcal{L}_{rec} + \lambda_2 \mathcal{L}_{SCL} +  \lambda_3 \mathcal{L}_{LOL}.
\end{equation}

FedDis has been extensively evaluated on several public brain MRI datasets including MSISBI~\citep{ghosh2019understanding}, OASIS~\citep{lamontagne2019oasis}, MSLUB~\citep{lesjak2018novel}, ADNI~\citep{rieke2020future} and BraTS~\citep{menze2014multimodal}.

\textbf{Tips \& Tricks.}~After training with healthy subjects, the shape auto-encoder is used to detect the brain anomalies. It assumes that the model cannot properly reconstruct the anomalies (\textit{e.g.}~tumors) as the anomalies are not seen during training. Hence, the anomalies are the parts where the reconstruction error is high. Although the reported results showcase the effectiveness of FedDis on detecting the anomalies, it may not work well when the shape encoder generalises well to reconstruct some tumors. Hence, some extra constraints to avoid such scenarios could be helpful to improve the robustness and performance of the model.

\section{What Can We Learn from Computer Vision?}

We are now well aware that learning disentangled representations requires supervision or design and learning biases. Using task prior knowledge to incorporate proper biases to learn the desired disentangled representations is key for disentanglement in both domains. Medical applications can use, for instance, building blocks (Sec.\ \ref{sec:bblocks}) originally designed for computer vision tasks. One can also draw inspiration from how prior knowledge on the vision tasks has motivated the specific biases used. Below, with some exemplar computer vision tasks, we discuss the connections between disentanglement in the computer vision and medical domains. 

\subsection{Image-to-image translation}
Image-to-image (I2I) translation aims to translate one image into another without changing the shape, \textit{i.e.}~content, which differ in a specific characteristic (\textit{e.g.}~style). A representative model is MUNIT~\citep{huang2017iccv} for which we provide details in Sec.~\ref{sec:applications}.

\textbf{Connections to medical.}~Image-to-image translation in computer vision motivated many medical applications as we detailed in Sec.~\ref{sec:applications}. In fact, several medical models are directly built based on MUNIT such as the ones in medical I2I translation~\citep{kang2019sashimi}, multi-modal and cross-modal segmentation~\citep{yang2020cross}, and registration~\citep{qin2019ipmi}. The parallels here of domain-invariant spatial content and the domain-specific style representation, relate to separating anatomy and modality representations in the corresponding medical applications. A major difference though is that typically in medical image translation we are particularly sensitive to maintaining identity when changing style. Several vision works show examples of day to night where content has changed slightly in the background. Such change will not be desired in medical tasks.

\subsection{Facial attribute transfer}
This task concerns the generation of a synthetic face that contains the target attribute, but without altering the subject identity (\textit{e.g.}~adding bangs to a subjects forehead). Most methods that focus on facial attribute transfer struggle with: a) transferring more than one attribute at a time, b) generating images based on exemplars, and c) achieving high-fidelity results. The first model to address the aforementioned challenges is ELEGANT~\citep{xiao2018eccv}, which encodes disentangled attribute representations of two exemplars in a vector latent space and performs attribute swapping. Apart from ELEGANT, \citet{lin2021tpami} propose a GAN model with a domain classifier to learn to transfer attributes between multiple domains. \citet{he2019tip} present a GAN that conditions the face generation of opposite samples (\text{e.g.}~smile, no smile) using one-hot attribute vectors. \citet{zhou2017bmvc} exploit cycle consistency to transfer attributes, with the limitation that the attributes should have approximately the same spatial location.

\textbf{Connections to medical.}~When transferring facial attributes, the subject identity should be preserved and only some attributes transferred. This transferral is desirable in several medical applications such as brain aging~\cite{xia2019consistent} and controllable synthesis~\cite{thermos2021miccai}, where the synthesised brain or heart images should contain the identity information of the original images but with different ages or pathology. ELEGANT preserves the identity information by only modifying the local part of the image. The medical models similarly modify the local anatomy parts but also apply the identity or consistency losses to the remaining parts of the image. We should note that most face models rely on pre-trained or pre-extracted strong priors to identify facial features. Such strong priors are rarely available in medical imaging. 

\subsection{Pose estimation}
For pose estimation, the human body constitutes a strong content prior that can be exploited to encode body structure in a spatial and semantic latent space, to be used for equivariant tasks that require body joint position. \citet{lorenz2019cvpr} propose to apply the equivariance and invariance losses to learn the equivariant (content) and invariant (style) representations and use this type of disentanglement for this challenging articulated body pose estimation task. \citet{esser2018cvpr} adopt the disentanglement of the human body pose from the corresponding appearance (style) information in the context of a dual-encoder VAE setting, where they use the body-related factors for human appearance transfer and synthesis~\citep{esser2019iccv}. 

\textbf{Connections to medical.}~Similar to the human body, human organs~\textit{e.g.} brain and heart, have strong anatomical structure priors, which can be similarly used for learning disentangled representations with equivariance and invariance properties. For example, similar to the invariance loss in \citet{lorenz2019cvpr}, \citet{bercea2021feddis} applies the shape consistency loss to encourage the shape embeddings of brain MRI images to be invariant to Gamma shifts. However, it is not always possible to assume such strong structural priors as diseases or abnormalities exist.


\section{Limitations, Opportunities and Open Challenges}
\label{sec:opportunities}

In this section, we identify three key limitations of existing DRL methods and discuss ideas and research directions for improvement. We also present opportunities as well as various challenges to be addressed by the community.

\subsection{New strategies for learning disentangled representations}

\textbf{Limitation.}\ Learning disentangled representations requires complex architectures and objective functions. As we saw in Sec.~\ref{sec:applications}, most approaches employ several loss functions and modules and, hence multiple hyperparameters. While flexibility is desirable, tuning complex systems can be difficult and it creates a barrier for further adoption of the disentanglement paradigm by the broader research community. Methods that require less hyperparameter tuning or techniques for automating this process or less complex approaches will be welcomed. Below, we discuss three possible strategies to learn disentangled representations in a simpler fashion.

\textbf{Integrating self-supervised and contrastive learning.}~ Fundamentally speaking most disentanglement approaches we reviewed here use a reconstruction approach. This may not be necessary. Recently, contrastive learning \citep{He2020cvpr,chen2020icml,chen2020neurips,grill2020bootstrap,zbontar2021barlow} has shown impressive performance for self-supervised representation learning. In particular, patch-wise contrastive learning \citep{park2020ceccv} has been successfully used as an auxiliary loss function for reinforcing disentanglement \citep{zhou2021chest,tomar2021miccai}. Additionally, \citet{mitrovic2021iclr} and \citet{vonkugelgen2021arxiv} developed an understanding of contrastive learning from a causal perspective and argue that it can be interpreted as CSD where the representation is focusing on learning only the content, whilst developing style invariance. Methods such as MOCO~\citep{He2020cvpr}, SimCLR~\citep{chen2020icml,chen2020neurips}, BYOL~\citep{grill2020bootstrap}, and the Barlow Twins~\citep{zbontar2021barlow} achieve this through augmentation and regularisation. \citet{wang2021selfsupervised} use contrastive learning for disentangling group invariant representations. \citet{ren2021learning} propose to discover the disentangled representations with contrasting learning at the post-hoc stage. \citet{pmlr-v139-zimmermann21a} have taking it a step further to suggest that contrastive learning under certain assumptions can indeed invert the data generating process.   While it is possible to learn representations that are robust (invariant) to specific interventions, it remains challenging to design augmentations and regularisations which are invariant to general interventions. 

\textbf{Intervention as a prior.}~\citet{caselles2019neurips} suggest that a symmetry-based understanding of disentanglement can only be achieved upon interaction with an environment. To illustrate this point, \citet{suter2019icml} propose a disentanglement metric based on interventional robustness. Moreover, statistical independence between latent variables might not hold for real-life settings where the generating factors are correlated~\citep{dittadi2021transfer, trauble2021disentangled}. With this intuition, \citet{Besserve2020Counterfactuals} provide a causal understanding of disentanglement in generative models based on interventions and counterfactuals. \citet{leeb2021interventional} propose a strategy for probing the latent space of VAEs by applying interventions. Their method allows quantification of the consistency of the representation with a chosen prior as well as finding holes in the latent manifold. These works pave a new path for using interventions as a prior for DRL.

\textbf{Compositionality as a prior.}~As reported in Sec.~\ref{sec:cs_disentanglement}, in current CSD models the content is vaguely defined as domain-invariant~\citep{huang2017iccv}, task-equivariant~\citep{lorenz2019cvpr} or even simply as spatial and binary~\citep{chartsias2019mia}. These definitions usually point to the task-driven model designs for learning the desired content, which are tailored to specific datasets or tasks. Enforcing compositionality could be the solution for learning generalisable and robust content representations in vision. This intuition is based on the compositional nature of the human cognition, which is robust for recognising new concepts by composing individual components~\citep{stone2017teaching}. Considering compositionality within disentanglement could be a fruitful direction.

\subsection{Disentanglement with additional properties}

\textbf{Limitation.}\ Generalisation on unseen data is the holy grail even in medical applications~\citep{sermesant2021nature}. Although disentangled representations should be general, Recent studies \citep{montero2020role, schott2021visual} found that disentanglement does not guarantee, for instance, combinatorial generalisation (understand and produce novel combinations of familiar elements). Another important limitation is learning disentangled representation from correlated data \citep{trauble2021icml}. As detailed in Sec.~\ref{sec:domain_shifts}, real data is not \textit{i.i.d.}~and bias exist due to domain shifts. In these cases, it has been shown that factorization-based inductive biases as described in Sec.\ \ref{sec:vae} are not enough to learn the true generating factors. These biases can have significant implications for domain generalisation and fairness (biased towards sensitive attributes).

\textbf{Domain generalisable disentangled representations.}  Domain generalisation is a setting which considers that no information from the target domain is available and that a model trained on multiple source domains needs to generalise well to the unseen target domain~\citep{li2018learning,Wang2021Unseen}. To address this, \citet{meng2021tmi} use task-specific representations and feature clustering to achieve domain invariance, and \citet{liu2021miccai} use meta-learning to explicitly improve domain invariance in disentangled representations. Concepts of causal representation learning \citep{scholkopf2021ieee} (Sec.\ \ref{sec:causality}) can help when defining and becoming robust to domain shifts when there are data biases \citep{arjovsky2019invariant,Krueger2021rex}. Recent work \citep{wang2021selfsupervised} disentangles group-invariant representations in a self-supervised setting using ideas from causal invariance \citep{arjovsky2019invariant}. Learning robust and generalisable representations, however, remains an open problem.

\textbf{Fair disentangled representations.}\ Fairness is an important concept in machine learning whenever an algorithm tends to be biased towards sensitive attributes such as race or gender \citep{puyol2021fairness, puyol2021fairnessmiccai}. Therefore, a fair model should be invariant to sensitive attributes. Developing fair algorithms is tightly related to domain generalisation as detailed in \citet{creager2021environment} and disentanglement provides a useful framework for dealing with these issues~\citep{locatello2019fairness, creager2019flexibly, sarhan2020fairness,Xianjing2021miccai}.

\subsection{Robust measurements of disentanglement}

\textbf{Limitation.}\ As analysed by~\citet{locatello2019icml}, most of the metrics reported in Sec.~\ref{sec:metrics} require ground truth for each latent factor or do not perform consistently for different tasks and datasets. Additionally, as experiments of \citet{trauble2021icml} show, most existing metrics struggle when measuring the disentanglement of models trained with data that include correlated factors of variation. 

\textbf{Metrics for real data.}\ Although a recent method considers to measure the disentanglement of hierarchically structured representations~\citep{dang2021evaluating}, robust disentanglement metrics that work well with real-world data (with any form and structure of generative factors) is still an open challenge. On the other hand, CSD disentanglement has attracted significant attention, with the exception of the metrics proposed by \citet{liu2020metrics}, the development of metrics that work with latents of diverse dimensionality is still an open problem. Such metrics can be further exploited to improve disentanglement itself in an iterative manner, as \citet{estermann2020robust} have done.

\section{Conclusion}
\label{sec:conclusion}
Overall, disentangled representation learning is a tool for introducing inductive biases (expert knowledge) into deep learning settings in order to simulate real-life scenarios with non-\textit{i.i.d.}~data. In this article, we have reviewed methods for implicitly or explicitly forcing representations to be invariant or equivariant to specific changes in the input data. We have emphasised building blocks for introducing disentanglement into a diverse set of tasks. In summary, disentanglement can be achieved with modifications in the model architecture (\textit{e.g.}~MUNIT, StyleGAN) and/or regularisation constraints (\textit{e.g.}~$\beta$-VAE). We highlight that disentanglement can be especially useful in low data regimes where biases are more relevant. By detailing limitations, opportunities and open challenges we hope to inspire the community to continue to investigate this extremely important area for learning better data representations. 

\section*{Acknowledgments}
This work was supported by the Royal Academy of Engineering and Canon Medical Research Europe, and partially supported by the Alan Turing Institute under the EPSRC grant EP/N510129/1. S.A.\ Tsaftaris acknowledges the support of Canon Medical and the Royal Academy of Engineering and the Research Chairs and Senior Research Fellowships scheme (grant RCSRF1819\textbackslash8\textbackslash25). We thank the participants of the DREAM tutorials for feedback. For the purpose of open access, the author has applied a Creative Commons Attribution (CC BY) licence to any Author Accepted Manuscript version arising from this submission.

\bibliographystyle{model2-names.bst}\biboptions{authoryear}
\balance
\bibliography{refs}

\end{document}

%% file: Tex/applications_summary_table.tex
\afterpage{%
    \clearpage
    \begin{table*}
        \centering 
        \caption{Summary of the surveyed medical imaging applications. * denotes the examplar models that we detail in Sec.~\ref{sec:applications}. 
        }
        \vspace{2mm}
\label{tab:medical_applications_summary}
\resizebox{1\textwidth}{!}{
\begin{tabular}{|cc|c|c|c|c|c|c|}
\hline
\multicolumn{2}{|c|}{\textbf{Application}} &
  \textbf{Model} &
  \textbf{Framework} &
  \textbf{Factors} &
  \textbf{Organ} &
  \textbf{Modality} &
  \textbf{Code} \\ \hline
\multicolumn{1}{|c|}{\multirow{31}{*}{Synthesis}} &
  \multirow{5}{*}{Disease decomposition} &
  \citet{xia2020mia} * &
  GAN &
  Normal, Abnormal &
  Brain &
  MRI &
  \cmark \\ \cline{3-8} 

\multicolumn{1}{|c|}{} &
   &
  \citet{kobayashi2021decomposing} &
  VAE + GAN &
  Normal, Abnormal  &
  Brain &
  MRI &
  \cmark \\ \cline{3-8}
 
  \multicolumn{1}{|c|}{} &
   &
  \citep{couronne2021longitudinal} &
  VAE &
  \begin{tabular}[c]{@{}c@{}}Disease Progression \\  Inter-patient variability\end{tabular}&
  Brain &
  MRI &
  \cmark \\ \cline{3-8} 

\multicolumn{1}{|c|}{} &
   &
  \citet{Tang2021mia} &
  VAE + GAN &
  Normal, Abnormal  &
  Lungs &
  X-ray &
  \xmark \\ \cline{2-8} 
  
\multicolumn{1}{|c|}{} &
  \multirow{6}{*}{I2I translation} &
  \citet{kang2019sashimi} &
  CSD &
  Content, Style &
  Eye &
  \begin{tabular}[c]{@{}c@{}}Fluorescein \\ Fundus\end{tabular} &
  \xmark \\ \cline{3-8} 

\multicolumn{1}{|c|}{} &
   &
  \citet{Zeju2019Anatomy} &
  GAN &
  Lung, Bones, Other &
  Chest &
  X-ray, CT &
  \cmark \\ \cline{3-8} 
\multicolumn{1}{|c|}{} &
   &
  \citet{Pfeiffer2019miccai} * &
  CSD &
  Content, Style &
  Liver, Abdomen &
  \begin{tabular}[c]{@{}c@{}}Laparoscopic\\ Images\end{tabular} &
  \cmark \\ \cline{3-8} 
\multicolumn{1}{|c|}{} &
   &
  \citet{fei2021deep} &
  CSD &
  Content, Style &
  Brain &
  \begin{tabular}[c]{@{}c@{}} MRI \end{tabular} &
  \xmark  \\ \cline{2-8} 
  
  \multicolumn{1}{|c|}{} & \multirow{4}{*}{Artefact reduction} &
  \citet{Liao2019tmi} * &
   GAN &
   Content, Artifact &
   Many &
   CBCT, CT & \cmark
   \\ \cline{3-8} 
\multicolumn{1}{|c|}{} & 
    &
  \citet{Huang2020miccai} &
   GAN &
   Content, Noise &
   Eye &
   OCT & \xmark
   \\ \cline{3-8} 
    \multicolumn{1}{|c|}{} & 
    &
  \citet{Niu2021Radiation} &
   GAN &
   Content, Artifact &
   Many l&
   CT & \xmark
\\ \cline{3-8} 
\multicolumn{1}{|c|}{} & 
    &
  \citet{tang2022generative} &
   GAN &
   Content, Artifact &
   Brain &
   MRI & \xmark
\\ \cline{2-8} 
   \multicolumn{1}{|c|}{} &
  \multirow{5}{*}{Harmonisation} &
  \citet{Wang2021miccaiflow} &
  VAE + NF &
  Sex, Age, Site, Image&
  Brain &
  MRI &
  \xmark \\ \cline{3-8} 
\multicolumn{1}{|c|}{} &
   &
  \citet{Dewey2020miccai} &
    CSD &
   Content, Style &
   Brain &
   MRI &
   \xmark \\ \cline{3-8} 
 \multicolumn{1}{|c|}{} &
   &
  \citet{Hongwei2021Uncertainty} &
   CSD &
   Content, Style &
   Brain &
   MRI &
   \xmark \\ \cline{3-8} 
\multicolumn{1}{|c|}{} &
   &
  \citet{Zuo2021neuroimage} * &
   CSD &
   Content, Style &
   Brain &
   MRI &
   \cmark \\ \cline{3-8} 
\multicolumn{1}{|c|}{} &
   &
  \citet{zuo2021informationbased} &
   CSD &
   Content, Style &
   Brain &
   MRI &
   \cmark \\ \cline{2-8} 
\multicolumn{1}{|c|}{} &
  \multirow{7}{*}{Controllable} &
  \citet{liu2018decompose} &
  CSD &
  Segmentation Meshes, Residue &
  Lung &
  CT &
  \xmark \\ \cline{3-8} 
 
\multicolumn{1}{|c|}{} &
   &
  \citet{Liu2020Manifold} &
  GAN &
  Content, Style &
  Chest &
  MRI, CT &
  \cmark \\ \cline{3-8} 
\multicolumn{1}{|c|}{} &
   &
  \citet{kelkar2022prior} &
  GAN &
  Content, Style &
  Brain &
  \begin{tabular}[c]{@{}c@{}} MRI \end{tabular} &
  \xmark \\ \cline{3-8} 
\multicolumn{1}{|c|}{} &
   &
  \citet{thermos2021miccai} * &
  CSD + GAN &
  Content, Style &
  Heart &
  MRI &
  \cmark \\ \cline{3-8}
  \multicolumn{1}{|c|}{} &
   &
  \citet{Hochberg2021Style} &
  GAN &
  Hierarchical Style &
  Liver &
  CT &
  \xmark \\ \cline{3-8} 
\multicolumn{1}{|c|}{} &
   &
  \citet{HAVAEI2021mia} &
  CSD + GAN &
  Content, Style&
  Skin, Lung &
  \begin{tabular}[c]{@{}c@{}}Cermatoscopic \\ Images, CT\end{tabular} &
  \xmark \\ \cline{2-8} 
\multicolumn{1}{|c|}{} &
  \multirow{5}{*}{Causal} &
  \citet{pawlowski2020neurips} * &
  VAE + NF &
  \begin{tabular}[c]{@{}c@{}} Sex, Age, \\ Brain Volume, Image \end{tabular} &
  Brain &
  MRI &
  \cmark \\ \cline{3-8} 
\multicolumn{1}{|c|}{} &
   &
  \citet{reinhold2021arxiv} &
  VAE + NF &
    \begin{tabular}[c]{@{}c@{}} Lesion, Ventricle and Brain volume,\\ Slice, Image, \\ Disability Score, Sex\end{tabular}&
  Brain &
  MRI &
  \cmark
   \\ \hline
   
\multicolumn{1}{|c|}{\multirow{28}{*}{Segmentation}} &
  \multirow{10}{*}{Single-modal} &
  \citet{chartsias2019mia} * &
  CSD & 
  Content, Style &
  \begin{tabular}[c]{@{}c@{}}Heart, Abdomen, \\ Spinal Cord \end{tabular} &
  MRI, CT &
  \cmark \\ \cline{3-8}
\multicolumn{1}{|c|}{} &
  &
  \citet{valvano2019temporal} &
  CSD & 
  Content, Style, Temporal &
  \begin{tabular}[c]{@{}c@{}}Heart\end{tabular} &
  MRI &
  \cmark \\ \cline{3-8}
\multicolumn{1}{|c|}{} &
   &
  \citet{jiang2020semi} &
   CSD &
   \begin{tabular}[c]{@{}c@{}}Content, Pathology, \\Modality  \end{tabular} &
   Heart &
   MRI, LGE  & 
   \cmark \\ \cline{3-8} 
\multicolumn{1}{|c|}{} &
  &
  \citet{liu2020mnms} &
  CSD & 
  Content, Style &
  \begin{tabular}[c]{@{}c@{}}Heart\end{tabular} &
  MRI &
  \cmark \\ \cline{3-8} 
\multicolumn{1}{|c|}{} &
   &
  \citet{shin2021unsupervised} &
   Auto-Encoder &
   Intensity, Non-intensity &
   Abdomen &
   CT  & 
   \cmark \\ \cline{3-8} 
\multicolumn{1}{|c|}{} &
  &
  \citet{liu2021miccai} &
  CSD & 
  Content, Style, Domain &
  \begin{tabular}[c]{@{}c@{}}Heart, Spinal Cord \end{tabular} &
  MRI &
  \cmark \\ \cline{3-8} 
\multicolumn{1}{|c|}{} &
   &
  \citet{memmel2021adversarial} &
   VAE + GAN &
   \begin{tabular}[c]{@{}c@{}}Content, Domain  \end{tabular} &
   Brain &
   MRI  & 
   \cmark \\ \cline{3-8} 
\multicolumn{1}{|c|}{} &
   &
  \citet{kalkhof2022disentanglement} &
   CSD &
   \begin{tabular}[c]{@{}c@{}}Content, Domain \end{tabular} &
   Bowel &
   CT  & 
   \xmark \\ \cline{2-8} 
\multicolumn{1}{|c|}{} &
  \multirow{18}{*}{Multi-modal} &
  \citet{yang2019unsupervised} &
  CSD &
  \begin{tabular}[c]{@{}c@{}}Content, Style \end{tabular} &
   Liver &
   MRI, CT & \xmark
   \\ \cline{3-8} 
\multicolumn{1}{|c|}{} &
   &
  \citet{chen2019robust} &
  CSD &
  \begin{tabular}[c]{@{}c@{}}Content, Style \end{tabular} &
   Brain &
   MRI & \xmark
   \\ \cline{3-8}
\multicolumn{1}{|c|}{} &
   &
  \citet{chen2019unsupervised} * &
  CSD &
  \begin{tabular}[c]{@{}c@{}}Content, Style \end{tabular} &
   Heart &
   MRI, LGE & \xmark
   \\ \cline{3-8} 
\multicolumn{1}{|c|}{} &
   &
  \citet{jiang2020unified} &
  CSD &
  \begin{tabular}[c]{@{}c@{}}Content, Style, Domain \end{tabular} &
   Abdomen &
   MRI, CT & \xmark
   \\ \cline{3-8} 
\multicolumn{1}{|c|}{} &
   &
  \citet{lyu2020joint} &
   CSD + GAN &
   Content, Style &
   Spine &
   CBCT, CT & \xmark
   \\ \cline{3-8} 
\multicolumn{1}{|c|}{} &
   &
  \citet{yang2020cross} &
  CSD &
  \begin{tabular}[c]{@{}c@{}}Content, Style \end{tabular} &
   Liver &
   MRI, CT & \xmark
   \\ \cline{3-8} 
\multicolumn{1}{|c|}{} &
   &
  \citet{xie2020mi} &
  Auto-Encoder, GAN &
  \begin{tabular}[c]{@{}c@{}} Content \end{tabular} &
   Eye, Colon &
   \begin{tabular}[c]{@{}c@{}}Colonoscopic Images, \\Fundus Images \end{tabular} &
   \xmark
   \\ \cline{3-8} 
 \multicolumn{1}{|c|}{} &
   &
  \citet{Runze2021Cross} &
  CSD + GAN &
  Content, Style &
  Heart &
  MRI &
  \xmark \\ \cline{3-8} 
\multicolumn{1}{|c|}{} &
   &
  \citet{ouyang2021representation} &
   CSD &
   Content, Style &
   Brain &
   MRI & \cmark
   \\ \cline{3-8} 
\multicolumn{1}{|c|}{} &
   &
  \citet{chartsias2019tmi} &
  CSD &
  \begin{tabular}[c]{@{}c@{}}Content, Style \end{tabular} &
   Heart, Abdomen&
   MRI, LGE & \cmark
   \\ \cline{3-8} 
\multicolumn{1}{|c|}{} &
   &
  \citet{pei2021disentangle} &
  CSD &
  \begin{tabular}[c]{@{}c@{}}Content, Style \end{tabular} &
   Heart &
   MRI, CT & \cmark
   \\ \cline{3-8} 
\multicolumn{1}{|c|}{} &
   &
  \citet{chen2021beyond} &
   CSD &
  \begin{tabular}[c]{@{}c@{}}Content, Style \end{tabular} &
   Heart &
   MRI, CT & \cmark
   \\ \cline{3-8}
\multicolumn{1}{|c|}{} &
   &
  \citet{chen2021diverse} &
  CSD &
  \begin{tabular}[c]{@{}c@{}} Content, Style \end{tabular} &
   Brain, Heart &
   MRI, CT & \xmark
   \\ \cline{3-8} 
\multicolumn{1}{|c|}{} &
   &
  \citet{ning2021new} &
  CSD &
  \begin{tabular}[c]{@{}c@{}} Content, Style, Domain \end{tabular} &
   Heart, Abdomen &
   MRI, CT & \xmark
   \\ \cline{3-8} 
\multicolumn{1}{|c|}{} &
   &
  \citet{wang2022cycmis} &
  CSD &
  \begin{tabular}[c]{@{}c@{}} Content, Style \end{tabular} &
   Heart &
   MRI, CT & \xmark
   \\ \cline{3-8}
\multicolumn{1}{|c|}{} &
   &
  \citet{liu2022multi} &
  CSD &
  \begin{tabular}[c]{@{}c@{}} Content, Style \end{tabular} &
   Nuclei  &
   Microscopy Images & \xmark
   \\ \hline
\multicolumn{2}{|c|}{\multirow{18}{*}{Classification}} &
  \citet{benvohen2019embc} &
   Classifier + GAN &
   Class Specific, Agnostic &
   Liver &
   CT & \xmark
   \\ \cline{3-8} 
\multicolumn{2}{|c|}{} &
  \citet{Gyawali2019Ensembling} &
   VAE &
   Class Specific, Agnostic &
   Chest &
   X-ray & \cmark
   \\ \cline{3-8}
\multicolumn{2}{|c|}{} &
  \citet{Zhao2019miccai} &
   VAE + Regressor &
   Age-specific, Agnostic &
   Brain &
   MRI & \cmark
   \\ \cline{3-8} 
  \multicolumn{2}{|c|}{} &

\citet{Puyol2020Resynchronisation} &
   VAE + Classifier &
   Different Biomarkers &
   Heart &
   MRI & \cmark
   \\ \cline{3-8}
\multicolumn{2}{|c|}{} &
  \citet{meng2021tmi} * &
   Classifier + Clustering &
   Categorical, Domain &
   Fetal &
   Fetal Ultrasound & \xmark
   \\ \cline{3-8}  
 \multicolumn{2}{|c|}{} &
  \citet{bass2020icam,bass2021icamreg} &
   CSD &
   Task Specific, Agnostic &
   Brain &
   MRI & \cmark
   \\ \cline{3-8}
 \multicolumn{2}{|c|}{} &
  \citet{Zou2021interpretation} &
   VAE &
   Sex, Rest &
   Hip &
   CT & \xmark
   \\ \cline{3-8} 
\multicolumn{2}{|c|}{} &
  \citet{zhao2021miccai} &
   VAE + Regressor &
   \begin{tabular}[c]{@{}c@{}} Age-specific, \\ Site-specific, Agnostic\end{tabular}
   &
   Bowel &
   Endoscopic Images & \xmark
   \\ \cline{3-8} 
\multicolumn{2}{|c|}{} &
  \citet{harada2021order} &
   GAN &
   \begin{tabular}[c]{@{}c@{}} Location-dependent, \\ Ulcerative Colitis-dependent\end{tabular}
   &
   Brain &
   MRI & \cmark
   \\ \cline{3-8} 
\multicolumn{2}{|c|}{} &
  \citet{wang2021miccai} &
  Auto-Encoder &
  \begin{tabular}[c]{@{}c@{}} Disease-related, \\ Disease-irrelevant\end{tabular}  &
   Breast&
   \begin{tabular}[c]{@{}c@{}}X-ray\end{tabular} 
   & \xmark
   \\ \cline{3-8}
\multicolumn{2}{|c|}{} &
  \citet{gravina2021dae} &
  Auto-Encoder + GAN &
  \begin{tabular}[c]{@{}c@{}} Shading, Albedo,\\ Deformation Field \end{tabular}  &
   Breast&
   \begin{tabular}[c]{@{}c@{}}MRI\end{tabular} 
   & \xmark
   \\ \cline{3-8}
\multicolumn{2}{|c|}{} &
  \citet{yang2021disentangled} &
  VAE &
  \begin{tabular}[c]{@{}c@{}} Time-invariant,\\ Time-variant \end{tabular}  &
   Brain &
   \begin{tabular}[c]{@{}c@{}}MRI\end{tabular} 
   & \xmark
   \\ \cline{3-8}
\multicolumn{2}{|c|}{} &
  \citet{zhou2022lung} &
  CSD + VAE &
  \begin{tabular}[c]{@{}c@{}} Structure, Texture\end{tabular}  &
   Lung &
   \begin{tabular}[c]{@{}c@{}}Chest Radiograph\end{tabular} 
   & \cmark
   \\ \hline
\multicolumn{2}{|c|}{\multirow{3}{*}{Registration}} &
  \citet{qin2019ipmi} * &
   CSD + GAN&
   Content, Style&
   Lung, Brain&
   MRI, CT & \xmark
   \\ \cline{3-8} 
\multicolumn{2}{|c|}{} &
     \citet{chartsias2019tmi} &
   CSD &
   Content, Style&
   Heart, Abdomen &
   MRI, LGE & \cmark
   \\ \cline{3-8} 
\multicolumn{2}{|c|}{} &
  \citet{maillard2022deep} &
   CSD &
   Content, Style&
   Brain&
   MRI & \xmark
   \\ \hline
\multicolumn{2}{|c|}{\multirow{2}{*}{Federated learning}} &
  \citet{li2020federated} & 
   CSD + GAN &
   Content, Style & 
   Heart & 
   CBCT, CT & \cmark
   \\ \cline{3-8}
\multicolumn{2}{|c|}{} &
  \citet{bercea2021feddis} * & 
   Auto-Encoder &
   Content, Style & 
   Brain & 
   MRI & \xmark
   \\ \hline
\end{tabular}
        } 
\end{table*}
    \clearpage
}